\pdfoutput=1
\documentclass{article}

\usepackage{microtype}
\usepackage{graphicx}
\usepackage{subcaption}
\usepackage{booktabs}
\usepackage{hyperref}

\usepackage[preprint]{icml2026}
\usepackage{amsmath}
\usepackage{amssymb}
\usepackage{mathtools}
\usepackage{algorithm}
\usepackage{algorithmic}
\usepackage[capitalize,noabbrev]{cleveref}
\usepackage{xspace}

\newcommand{\CQwMatch}{148}
\newcommand{\ablCPAms}{13.7}
\newcommand{\ablCPAsg}{6.1}
\newcommand{\ablCPLms}{14.2}
\newcommand{\ablCPLsg}{6.3}

\newcommand{\ablPsrAms}{12.1}
\newcommand{\ablPsrAsg}{3.3}
\newcommand{\ablPsrLms}{12.8}

\newcommand{\ablPsrLsg}{3.8}

\newcommand{\arCopyEffGSM}{0.32}
\newcommand{\arCopyEffHE}{0.31}

\newcommand{\arQwFloorGsm}{22.9}
\newcommand{\arQwFloorHe}{23.0}

\newcommand{\arQwHalfRng}{24.0--25.0}
\newcommand{\arQwMatchRng}{24.1--24.8}
\newcommand{\ciQwCP}{0.8}
\newcommand{\ciQwPsr}{0.7}

\newcommand{\dQwCPHella}{$-$0.85}

\newcommand{\dQwCPMmlu}{$-$1.27}

\newcommand{\dQwPairHe}{$-$2.4}
\newcommand{\dQwPairHella}{+0.57}
\newcommand{\dQwPairIfe}{$-$2.2}

\newcommand{\dQwPairMmlu}{+0.85}

\newcommand{\dQwPsrMmlu}{$-$0.43}
\newcommand{\detPhysFourA}{8.63}
\newcommand{\detPhysFourB}{8.86}
\newcommand{\detPhysOneA}{8.76}
\newcommand{\detPhysOneB}{9.10}
\newcommand{\detPhysRngA}{0.054}
\newcommand{\detPhysRngB}{0.027}
\newcommand{\dispQwCP}{78--126}
\newcommand{\dispQwPsr}{53--67}
\newcommand{\ebCImax}{4.8}
\newcommand{\ebCImin}{3.2}

\newcommand{\ebDiscExactGsm}{21}
\newcommand{\ebExactGsmeng}{32.8}
\newcommand{\ebExactGsmhar}{33.2}

\newcommand{\ebNomDiff}{3.7}

\newcommand{\ebNtests}{20}

\newcommand{\expCopyCPtwoA}{0.70}

\newcommand{\expCopyExactLruA}{0.59}
\newcommand{\expCopyFastA}{0.28}

\newcommand{\fgCPGsmcp}{5.2}

\newcommand{\fgCPGsmpn}{23.4}

\newcommand{\fgCPHecp}{1.4}

\newcommand{\fgCPHepn}{14.6}

\newcommand{\fgCPIfecp}{7.5}

\newcommand{\fgCPIfepn}{29.5}

\newcommand{\fgFastGsmcp}{10.9}

\newcommand{\fgFastGsmpn}{14.0}

\newcommand{\fgFastHecp}{6.9}

\newcommand{\fgFastHepn}{11.2}

\newcommand{\fgFastIfecp}{12.0}

\newcommand{\fgFastIfepn}{18.3}

\newcommand{\fgFullDiff}{6.4}
\newcommand{\fgNGsm}{20}
\newcommand{\fgNHe}{41}
\newcommand{\fgNIfe}{16}
\newcommand{\fgNprompts}{77}
\newcommand{\fgNrounds}{3}

\newcommand{\fgNtests}{9}

\newcommand{\fgPrLenCPIfe}{771}

\newcommand{\fgPrLenPsrIfe}{615}

\newcommand{\fgPrNQGsm}{20}

\newcommand{\fgPrNmax}{41}
\newcommand{\fgPrNmin}{16}

\newcommand{\fgPrWinQGsm}{13}

\newcommand{\fgQwCPGsm}{24.8}
\newcommand{\fgQwCPHe}{24.3}
\newcommand{\fgQwCPcpGsm}{9.2}
\newcommand{\fgQwCPcpHe}{6.4}

\newcommand{\fgQwPsrGsm}{25.8}
\newcommand{\fgQwPsrHe}{25.7}
\newcommand{\fgQwPsrcpGsm}{24.7}
\newcommand{\fgQwPsrcpHe}{23.9}

\newcommand{\fgQwSlowRng}{4--6}
\newcommand{\fgRedGsm}{12}
\newcommand{\fgRedGsmci}{8--16}

\newcommand{\fgRedIfe}{18}
\newcommand{\fgRedIfeci}{11--25}
\newcommand{\fgRedPrGsm}{18}
\newcommand{\fgRedPrHe}{7}
\newcommand{\fgRedPrIfe}{28}

\newcommand{\fgRedqGsm}{28}

\newcommand{\fgRedqHe}{19}

\newcommand{\fgRedqIfe}{24}

\newcommand{\fgRedqRng}{19--28}

\newcommand{\fgSlowHe}{4}
\newcommand{\fgSlowHeci}{3--5}

\newcommand{\fgSpdExactRng}{1.7--2.3}
\newcommand{\insLaterAA}{2.5}
\newcommand{\insLaterAB}{5.1}
\newcommand{\insLaterLruA}{7.3}
\newcommand{\insLaterLruB}{10.4}
\newcommand{\klMoonCPA}{0.028}
\newcommand{\klMoonCPAHalf}{0.020}
\newcommand{\klMoonCPAQuarter}{0.038}
\newcommand{\klMoonCPfour}{0.072}
\newcommand{\klMoonCPsix}{0.144}
\newcommand{\klMoonPsr}{0.021}
\newcommand{\klMoonPsrHalf}{0.014}
\newcommand{\klMoonPsrHi}{0.029}
\newcommand{\klMoonPsrQuarter}{0.030}
\newcommand{\klMoonPsrSix}{0.078}
\newcommand{\klQwCP}{0.051}
\newcommand{\klQwCPJtwo}{0.047}
\newcommand{\klQwCPthree}{0.119}
\newcommand{\klQwPsr}{0.029}
\newcommand{\klQwPsrTwo}{0.036}
\newcommand{\klQwRatio}{0.56}
\newcommand{\lateFastA}{9.9}
\newcommand{\lateSpecA}{15.3}

\newcommand{\meanExFastA}{0.0}
\newcommand{\meanExFastAci}{0.9}
\newcommand{\meanExFastAnohe}{$-$0.8}
\newcommand{\meanExFastAnoheci}{0.8}

\newcommand{\meanExMoonCPA}{$-$1.37}

\newcommand{\meanExMoonCPfour}{$-$2.03}

\newcommand{\meanExMoonPsr}{$-$0.48}
\newcommand{\meanExMoonPsrHi}{$-$1.03}

\newcommand{\meanExMoonPsrci}{0.8}

\newcommand{\meanExQualAnohe}{$-$0.1}
\newcommand{\meanExQualAnoheci}{0.7}

\newcommand{\meanExSpec}{$-$0.4}
\newcommand{\meanExSpecci}{0.8}

\newcommand{\memQwArHalf}{19.7}
\newcommand{\memQwArMatch}{22.3}
\newcommand{\memQwSd}{22.4}
\newcommand{\moonCIpct}{17}
\newcommand{\moonCIpsrpct}{8}
\newcommand{\msArFloorGSM}{6.5}
\newcommand{\msArFloorHE}{6.1}
\newcommand{\msArPsrGSM}{11.3}
\newcommand{\msArPsrHE}{9.2}
\newcommand{\msCPAnopfA}{13.8}
\newcommand{\msCPAthA}{16.3}
\newcommand{\msCPthreeA}{11.1}

\newcommand{\msCPtwoA}{14.3}
\newcommand{\msCPtwoB}{27.5}
\newcommand{\msExactBestA}{20.5}
\newcommand{\msExactBestB}{37.8}
\newcommand{\msExactLruA}{24.7}
\newcommand{\msExactLruB}{42.5}
\newcommand{\msFastA}{10.7}
\newcommand{\msFastB}{19.2}
\newcommand{\msFloor}{6.8}

\newcommand{\msJtwoMaxA}{11.2}
\newcommand{\msJtwoMaxB}{15.7}
\newcommand{\msJtwoThreeA}{12.1}

\newcommand{\msJtwoTwoA}{14.4}

\newcommand{\msMoonCPA}{25.9}
\newcommand{\msMoonCPAHalf}{19.4}
\newcommand{\msMoonCPAQuarter}{36.4}
\newcommand{\msMoonCPfour}{21.5}
\newcommand{\msMoonCPsix}{18.6}
\newcommand{\msMoonExact}{43.6}
\newcommand{\msMoonExactHalf}{28.9}
\newcommand{\msMoonExactQuarter}{60.1}
\newcommand{\msMoonPrefetch}{34.6}
\newcommand{\msMoonPsr}{22.6}
\newcommand{\msMoonPsrHalf}{18.2}

\newcommand{\msMoonPsrQuarter}{29.6}
\newcommand{\msMoonPsrSix}{17.8}

\newcommand{\msPsrFifA}{13.1}
\newcommand{\msPsrNtwoA}{14.5}
\newcommand{\msQualA}{12.0}
\newcommand{\msQualB}{24.7}
\newcommand{\msQwCP}{23.6}
\newcommand{\msQwCPJtwo}{23.5}
\newcommand{\msQwCPthree}{22.7}
\newcommand{\msQwExact}{30.0}
\newcommand{\msQwExactHalf}{23.7}
\newcommand{\msQwExactMid}{26.0}

\newcommand{\msQwFloor}{22.0}

\newcommand{\msQwPsr}{24.1}

\newcommand{\msQwPsrTwo}{23.5}
\newcommand{\msQwRngHalf}{22.0--22.8}
\newcommand{\msQwRngMid}{22.2--23.5}
\newcommand{\msQwRngQuarter}{22.7--24.9}
\newcommand{\msQwSpread}{3}
\newcommand{\msSpecA}{20.0}
\newcommand{\msSpecB}{38.1}
\newcommand{\msTdepA}{13.9}
\newcommand{\msTdepLoA}{14.7}
\newcommand{\msUnbA}{17.0}

\newcommand{\msUnbLruA}{18.7}

\newcommand{\nRep}{3}
\newcommand{\nRepMoon}{3}
\newcommand{\nRepQw}{3}
\newcommand{\nTestsBench}{90}
\newcommand{\nTestsMoon}{54}
\newcommand{\nTestsPairs}{81}
\newcommand{\nTestsQwen}{21}
\newcommand{\pQwPairGenMin}{0.13}
\newcommand{\pairIsoLatA}{+1.7}
\newcommand{\pairIsoLatAci}{1.0}

\newcommand{\pairIsoLatAnohe}{+0.9}

\newcommand{\pairIsoLatB}{+2.2}
\newcommand{\pairIsoLatBci}{1.0}

\newcommand{\pairIsoLatBnohe}{+1.8}

\newcommand{\pairIsoQualA}{+0.1}
\newcommand{\pairIsoQualAci}{0.9}
\newcommand{\pairIsoQualAlo}{$-$0.8}

\newcommand{\pairIsoQualAnohelo}{$-$1.0}
\newcommand{\pairIsoQualAq}{+0.1}
\newcommand{\pairIsoQualAqci}{0.8}
\newcommand{\pairIsoQualAqlo}{$-$0.8}

\newcommand{\pairIsoQualAqnohelo}{$-$0.3}
\newcommand{\pairIsoQualB}{+0.5}
\newcommand{\pairIsoQualBci}{0.8}
\newcommand{\pairIsoQualBlo}{$-$0.3}

\newcommand{\pairMoonIsoLat}{+1.6}
\newcommand{\pairMoonIsoLatci}{0.9}
\newcommand{\pairMoonIsoQual}{+0.9}
\newcommand{\pairMoonIsoQualci}{0.8}
\newcommand{\pairQw}{$-$0.4}
\newcommand{\pairQwci}{1.0}
\newcommand{\pairSpecA}{+0.4}
\newcommand{\pairSpecAci}{0.9}

\newcommand{\pairSpecB}{0.0}
\newcommand{\pairSpecBci}{1.0}

\newcommand{\pfAllFastA}{11.7}

\newcommand{\pfOnTimeFastA}{0.5}

\newcommand{\pfUnusedFastA}{1.3}

\newcommand{\pfUsedFastA}{10.4}

\newcommand{\pplPsrFifA}{8.60}
\newcommand{\pplPsrNtwoA}{8.55}
\newcommand{\pplTdepA}{8.65}
\newcommand{\pplTdepLoA}{8.56}
\newcommand{\ratCPA}{1.33}
\newcommand{\ratCPAhi}{1.38}
\newcommand{\ratCPAlo}{1.28}

\newcommand{\ratCPAq}{1.19}
\newcommand{\ratCPAqhi}{1.23}
\newcommand{\ratCPAqlo}{1.15}

\newcommand{\ratExactBestA}{1.92}
\newcommand{\ratExactBestAhi}{2.05}
\newcommand{\ratExactBestAlo}{1.80}

\newcommand{\ratExactLruA}{2.31}
\newcommand{\ratExactLruAhi}{2.44}
\newcommand{\ratExactLruAlo}{2.19}
\newcommand{\ratExactLruAp}{$<$0.001}

\newcommand{\ratIsoA}{1.04}

\newcommand{\ratIsoAp}{0.05}

\newcommand{\redCPA}{25}
\newcommand{\redCPB}{30}

\newcommand{\redMoonCPA}{13}
\newcommand{\redMoonHi}{6}
\newcommand{\runAvgMaxDiff}{1.3}
\newcommand{\scQwArExactGsm}{78.1}

\newcommand{\scQwArFastGsm}{59.4}
\newcommand{\scQwSdConfGsm}{75.0}

\newcommand{\sdBEHE}{24}
\newcommand{\sdBEhidGSM}{15}
\newcommand{\sdBEhidHE}{35}
\newcommand{\sdBEhidStepsGSM}{2.4}
\newcommand{\sdBEhidStepsHE}{5.6}

\newcommand{\sdBEstepsHE}{3.9}
\newcommand{\sdBytesConfGSM}{1.03}
\newcommand{\sdBytesConfHE}{0.56}
\newcommand{\sdBytesPosGSM}{1.22}
\newcommand{\sdBytesPosHE}{0.67}

\newcommand{\sdCopyMsGSM}{0.60}
\newcommand{\sdCopyMsHE}{0.57}
\newcommand{\sdDraftGSM}{61}
\newcommand{\sdDraftHE}{56}
\newcommand{\sdDraftStep}{3.5--3.8}
\newcommand{\sdDraftTokGSM}{13.8}
\newcommand{\sdDraftTokHE}{8.9}

\newcommand{\sdFreeGSM}{1.45}
\newcommand{\sdFreeHE}{1.15}
\newcommand{\sdQwConfBGsm}{11.1}
\newcommand{\sdQwConfBHe}{8.3}

\newcommand{\sdQwConfGsm}{11.2}
\newcommand{\sdQwConfHe}{8.4}

\newcommand{\sdQwDraftBlkGsm}{10}

\newcommand{\sdQwDraftGsm}{1.3}
\newcommand{\sdQwDraftHe}{0.8}
\newcommand{\sdQwExactGsm}{23.0}
\newcommand{\sdQwExactHe}{15.8}

\newcommand{\sdQwFreeGsm}{7.0}
\newcommand{\sdQwFreeHe}{5.1}
\newcommand{\sdQwTauGsm}{7.7}
\newcommand{\sdQwTauHe}{13.0}
\newcommand{\sdRatioGSM}{1.99}
\newcommand{\sdRatioHE}{1.41}

\newcommand{\sdVerCompGSM}{11.7}
\newcommand{\sdVerCompHE}{10.6}
\newcommand{\sdVerRatioGSM}{1.10}
\newcommand{\sdVerRatioHE}{0.58}

\newcommand{\sgCPthreeA}{18.2}

\newcommand{\sgFastA}{5.4}

\newcommand{\sgJtwoHalfB}{34.5}
\newcommand{\sgJtwoMaxA}{14.2}
\newcommand{\sgJtwoMaxB}{38.0}
\newcommand{\sgJtwoThreeA}{10.2}

\newcommand{\sgJtwoTwoA}{5.8}

\newcommand{\sgSpec}{5.4}
\newcommand{\sgUnbA}{1.4}
\newcommand{\sgUnbLruA}{1.7}
\newcommand{\spdExactBestA}{1.9}
\newcommand{\spdExactBestB}{2.0}
\newcommand{\spdExactLruA}{2.3}
\newcommand{\spdExactLruB}{2.2}
\newcommand{\spdExactLruRng}{2.2--2.3}
\newcommand{\spdMoonExact}{1.9}
\newcommand{\spdQwExactHalf}{1.05}
\newcommand{\spdQwExactMid}{1.12}
\newcommand{\spdQwExactQuarter}{1.25}

\newcommand{\spdQwSdBootGsmci}{2.1--2.2}

\newcommand{\spdQwSdBootHeci}{2.8--3.0}

\newcommand{\spdQwSdCPRng}{2.1--2.9}

\newcommand{\spdQwSdExactRng}{2.5--3.3}
\newcommand{\spdQwSdExactVerGsm}{1.04}
\newcommand{\spdQwSdExactVerHe}{1.52}

\newcommand{\spdQwSdFloorRng}{2.0--2.7}
\newcommand{\spdSpecA}{1.9}
\newcommand{\spdSpecRng}{1.9--2.0}
\newcommand{\stallCPtwoA}{10.5}
\newcommand{\stallFastA}{2.2}
\newcommand{\stallSpecA}{13.3}
\newcommand{\tailCPtwoA}{30}

\newcommand{\tailExactLruA}{52}

\newcommand{\tailFastA}{19}

\newcommand{\tailMoonCPfour}{43}
\newcommand{\tailMoonExact}{96}
\newcommand{\tailMoonPsr}{47}

\newcommand{\tailSpecA}{43}

\newcommand{\trBBH}{+1.7}

\newcommand{\trBBHexUn}{$-$2.1}
\newcommand{\trBBHexUnp}{.01}
\newcommand{\trBBHp}{.027}
\newcommand{\trBonusMean}{+0.6}
\newcommand{\trBonusMeanci}{0.9}
\newcommand{\trHella}{$-$0.8}
\newcommand{\trMean}{+0.3}

\newcommand{\trMeanci}{0.8}
\newcommand{\valQwDD}{$-$0.002}
\newcommand{\valQwDDci}{0.006}
\newcommand{\valQwEff}{0.008}
\newcommand{\valQwPpl}{0.12}
\newcommand{\xferExactA}{30.5}
\newcommand{\xferFastA}{13.8}
\newcommand{\xferMoonPsr}{18.5}
\newcommand{\xferMoonPsrWide}{28.0}

\newcommand{\xferQwExact}{54}
\newcommand{\xferQwFloor}{0.7}

\newcommand{\xferSpecA}{29.1}

\newcommand{\psr}{PSR\xspace}
\newcommand{\cp}{Cache-Prior\xspace}

\icmltitlerunning{Prefetch-Steered Routing for Offloaded MoE Decoding}

\begin{document}

\twocolumn[
  \icmltitle{Prefetch-Steered Routing: Deterministic Cache-Aware Expert Selection\\
    for Offloaded Mixture-of-Experts Decoding}

  \begin{icmlauthorlist}
    \icmlauthor{Siming HUANG}{hkustgz}
  \end{icmlauthorlist}
  \icmlaffiliation{hkustgz}{The Hong Kong University of Science and Technology (Guangzhou), Guangzhou, China}
  \icmlcorrespondingauthor{Siming HUANG}{shuang886@connect.hkust-gz.edu.cn}
  \icmlkeywords{Mixture of Experts, Offloading, Expert Caching, Prefetching, LLM Inference}
  \vskip 0.3in
]

\printAffiliationsAndNotice{}

\begin{abstract}
When a Mixture-of-Experts (MoE) model does not fit in GPU memory, batch-1 decoding is dominated by copying expert weights from host memory. Cache-aware routing reduces these copies by biasing the router toward resident experts, but combined with asynchronous prefetching it makes the model's outputs depend on transfer timing, so a harness that does not simulate that timing no longer evaluates the deployed model. We propose \emph{prefetch-steered routing} (\psr): routing conditions on a deterministic \emph{logical} cache, into which at most $n$ experts predicted by the next layer's router are inserted per layer and token; physical copies affect only latency. On OLMoE-1B-7B with 50\% and 25\% of experts resident on an RTX~4090, \psr decodes \spdExactLruRng$\times$ faster than exact routing with LRU and \redCPA--\redCPB\% faster than the \cp operating point, with mean benchmark differences of \pairIsoQualA\ and \pairIsoQualB\ points (95\% lower bounds \pairIsoQualAlo\ and \pairIsoQualBlo). In greedy generation its lead over \cp depends on the workload and reverses on HumanEval at 50\% residency. On Moonlight-16B-A3B, \psr is \spdMoonExact$\times$ faster than exact routing with LRU at 37.5\% residency and, at the latency of \cp, significantly better on three of eight benchmarks. On Qwen3.6-35B-A3B, where copies are cheap relative to computation, \psr keeps its advantage over \cp on the likelihood benchmarks but not its speed. Distillation-trained gates did not help. Speculative decoding with block-level expert selection was slower than \psr with a 1B autoregressive drafter on OLMoE, but with a block drafter on Qwen3.6 it was \spdQwSdCPRng$\times$ faster than any autoregressive configuration with at most its GPU memory on the 64 prompts we timed.
\end{abstract}

\section{Introduction}
\label{sec:intro}

Sparse Mixture-of-Experts (MoE) layers~\citep{shazeer2017outrageously,fedus2022switch} decouple a model's parameter count from its per-token compute. This makes MoE models attractive for local, single-user inference, but their total weight often exceeds the memory of a consumer GPU. The standard remedy is expert offloading: non-expert weights stay on the GPU, experts live in host memory, and a per-layer GPU cache holds a subset of them~\citep{eliseev2023fast,xue2024moeinfinity}. At batch size one, each decoded token activates $K$ experts in every layer, and every cache miss is a PCIe copy on the critical path. Prefetching hides part of this latency~\citep{hwang2024pregated,song2024promoe,shen2025expertflow}, but when the router's choices are hard to predict and the link is close to saturated, transfers still dominate.

A second family of methods changes \emph{which} experts run. Cache-aware routing adds a bonus to the router logits of resident experts~\citep{skliar2025cache,bambhaniya2024eras}; expert substitution replaces missing experts with resident ones~\citep{zhu2025smoe}; speculative execution runs the experts predicted ahead of time instead of the router's choices~\citep{madan2026speculating,li2026commitmoe}. Bias and substitution trade a small deviation from the original model for large reductions in transfers; speculative execution instead moves transfers earlier. SpecMD~\citep{hoang2026specmd} studies how such routing, prefetching, eviction and miss-handling policies interact, including the routing bonus together with cross-layer prefetching. Two questions remain open.

\textbf{What is the model being evaluated?} Once routing depends on the cache, the cache becomes part of the model's semantics. If the cache is filled by asynchronous prefetches, the experts that are resident when the router runs depend on PCIe timing, contention, and scheduler parameters. The same prompt can then produce different outputs on different machines or under different load, and unless the evaluation harness reproduces the engine's timing, the quality it measures belongs to a different function from the one deployed. We show that this is a measurable effect, not a theoretical one (\cref{sec:semantics}). \citet{skliar2025cache} note that their algorithm is deterministic, which holds because their cache has no asynchronous prefetching. In SpecMD the bonus applies to the GPU-resident set, which an asynchronous prefetch thread fills, and in SMoE and MoE-CORE substitution uses whatever the prefetcher has made resident; this is our reading of their designs, and none of the three discusses the issue~\citep{hoang2026specmd,zhu2025smoe,yang2026moecore}. Speculating Experts and CommitMoE avoid it by executing predictions that do not depend on the cache, at the cost of executing every misprediction.

\textbf{How should routing bias and prefetching be combined?} SpecMD reports that the routing bonus accounts for 10--20\% of the speedup in some of its configurations. We find that \emph{how} the two are combined matters. Prefetching based on token history or expert popularity gives exact routing little in our engine and clearly \emph{hurts} under cache-aware routing: the bias already keeps routing on resident experts, so most prefetches are wasted and delay demand copies. Prefetching the experts that the \emph{next layer's router} predicts from the current hidden state behaves differently: the routing bonus then applies to experts the router was likely to choose anyway, so the bias steers routing onto in-flight experts rather than away from the router's preference.

We turn these observations into a design, \emph{prefetch-steered routing} (\psr; \cref{sec:design}):
\begin{itemize}
  \item Routing conditions only on a \textbf{logical cache}, a deterministic function of the token history and the policy. Physical copies are scheduled separately and change latency, never outputs. The evaluation harness and the deployed engine therefore implement the same routing policy; they differ only in floating-point numerics, which under greedy decoding change individual outputs but, in benchmarks generated by the engine itself, not the differences between configurations beyond the resolution of that check, about four points per task (\cref{sec:engine-quality}).
  \item A deterministic budget decides \textbf{which predicted experts receive the routing bonus}: at each layer and token, at most $n$ (1 or 2) of the next layer's predicted experts are inserted into the logical cache when they are predicted, independently of when their copies complete. A small budget keeps the added copies within what the link can carry; unbounded insertion gives the best quality but commits more transfers than the link can overlap.
  \item A score-based \textbf{eviction priority} (exponential moving averages of the router's choices plus a token prior) replaces LRU, as router-informed replacement policies do in systems with exact routing~\citep{zhong2025hybrimoe,yang2026moecore}.
\end{itemize}

We implement \psr in a batch-1 offloading engine with CUDA graphs and a fused MoE kernel, and in a HuggingFace evaluation harness that runs the identical logical cache model (\cref{sec:impl}). On OLMoE-1B-7B-Instruct~\citep{muennighoff2025olmoe} with 50\% and 25\% of experts resident on an RTX~4090 (\cref{sec:eval}):
\begin{itemize}
  \item \psr decodes at \msFastA\ ms/token at 50\% residency, \spdExactLruA$\times$ faster than exact routing with LRU (\msExactLruA\ ms) and \spdExactBestA$\times$ faster than the best exact-routing cache and prefetch configuration we found (\msExactBestA\ ms). At 25\% residency it runs at \msFastB\ ms/token (\spdExactLruB$\times$ and \spdExactBestB$\times$).
  \item Against the operating point of \cp routing~\citep{skliar2025cache} ($\lambda=0.2$), \psr has \redCPA\% (50\%) and \redCPB\% (25\%) lower latency; the mean benchmark difference is \pairIsoQualA\ and \pairIsoQualB\ points, with 95\% lower bounds of \pairIsoQualAlo\ and \pairIsoQualBlo. At matched latency it is ahead by \pairIsoLatA\ $\pm$\pairIsoLatAci\ and \pairIsoLatB\ $\pm$\pairIsoLatBci\ points; after Holm's correction it is better on HellaSwag (both sizes) and MMLU (25\%) and worse on no task. These latencies are teacher-forced; in greedy generation on the benchmark prompts at 50\% residency, the latency gap to \cp is \fgRedGsm\% on GSM8K and \fgRedIfe\% on IFEval, and \cp is faster on HumanEval; at 25\% residency \psr stays \fgRedqRng\% faster on all three (\cref{sec:engine-quality}).
  \item Speculative execution of predicted experts~\citep{madan2026speculating} is \spdSpecRng$\times$ slower than \psr and not significantly better on any benchmark, because it moves transfers earlier without reducing them.
  \item On Moonlight-16B-A3B~\citep{liu2025muon}, a DeepSeek-V3-style model~\citep{deepseekai2024v3} that must be offloaded on this GPU, \psr transfers without retuning (\cref{sec:moonlight}). At every cache size we ran, the latency--divergence frontier consists of lossless exact routing and \psr configurations. With 37.5\% of experts resident it runs at \msMoonPsr\ ms/token, \spdMoonExact$\times$ faster than exact routing with LRU. At the same latency as \cp it is ahead by \pairMoonIsoLat\ $\pm$\pairMoonIsoLatci\ points on average over eight benchmarks, significantly after correction on three of them.
  \item On Qwen3.6-35B-A3B~\citep{qwen36_35b_a3b} in FP8, with 256 experts per layer and expert copies a quarter the size of OLMoE's, decoding is bounded by computation (\cref{sec:qwen}). Every cache-aware configuration runs within \msQwSpread\ ms/token of the fastest, and \psr is only \spdQwExactQuarter$\times$ faster than exact routing with LRU at 25\% residency. At the same teacher-forced latency as \cp it has \klQwRatio\ times its KL and is significantly better on MMLU and HellaSwag, with no significant difference on the generative tasks; in free generation \cp is \fgQwSlowRng\% faster. \psr's latency advantage therefore requires copies to be a large share of the time per token.
\end{itemize}
Quality is measured with per-question paired tests on MMLU, ARC-Challenge, HellaSwag, GSM8K, BBH (chain-of-thought), IFEval, HumanEval and MBPP (\cref{sec:eval-quality}). We also report a negative and a conditional result. Training the gate (distillation to the frozen model with learned switching margins and LoRA adapters, and the same for \psr's own router) does not improve benchmark quality over training-free cache-aware routing at the same transfer budget (\cref{sec:training}). Speculative decoding with block-level expert selection~\citep{liang2026acceptmoe} reduces the copies per token below \psr's when drafts are accepted in long runs, but with a 1B autoregressive drafter on OLMoE it is slower than \psr on both tasks we timed. With a block drafter on Qwen3.6 it is \spdQwSdCPRng$\times$ faster than the fastest autoregressive configuration with at most its GPU memory on the 64 prompts we timed, whose scores detect only large losses (\cref{sec:sd}).

\textbf{Scope.} We study three models. OLMoE fits in the 24\,GB of an RTX~4090 in bf16, so for it we emulate memory pressure by capping the per-layer expert cache; Moonlight's 31\,GB and Qwen3.6's 37\,GB (FP8) do not fit, so their offloading is real. On Qwen3.6 we ran six of the eight benchmarks. We do not compare against complete external systems such as llama.cpp, MoE-Infinity or CPU-execution systems~\citep{kamahori2024fiddler,chen2025ktransformers}; all baselines run in our engines. \Cref{sec:limits} discusses these limits.

\section{Background}
\label{sec:background}

\textbf{MoE decoding with offloading.} An MoE layer $l$ has $E$ experts and a router that produces logits $z_{t,l}\in\mathbb{R}^E$ for token $t$. Exact routing executes the top-$K$ experts by $z_{t,l}$ and combines their outputs with softmax weights. We consider batch-1 decoding with all non-expert weights on the GPU and a per-layer cache of $C<E$ expert slots; the remaining experts sit in pinned host memory. A selected expert that is not resident must be copied before the layer can finish. Copies run on a separate stream and can overlap with computation of earlier layers if they are issued early enough.

\textbf{Cache-aware routing.} \cp routing~\citep{skliar2025cache} adds a bonus to the logits of resident experts before the top-$K$ selection:
\begin{equation}
  \tilde z_{t,l} = z_{t,l} + \lambda\,\bar\Delta_l\, m_{t,l},
  \label{eq:cp}
\end{equation}
where $m_{t,l}\in\{0,1\}^E$ marks resident experts, $\bar\Delta_l$ is the average range $\max z - \min z$ of the layer's logits (estimated once on held-out text), and $\lambda$ sets the strength. The selected experts are combined with the softmax weights of the \emph{unmodified} logits. With an LRU cache and no prefetching, the cache state is a deterministic function of past selections, so the routed model is a well-defined (if history-dependent) function of the input.

\textbf{Cross-layer prediction.} The router of layer $l$ applied to the hidden state entering layer $l-1$'s MoE block predicts layer $l$'s experts well, because the residual stream changes little between consecutive layers~\citep{eliseev2023fast,zhong2024adapmoe,madan2026speculating}; learned predictors refine this~\citep{hwang2024pregated,song2024promoe}. We call this ranking the \emph{pre-gate} ranking $g_{t,l}$. It is available one layer ahead, about 0.4\,ms at our layer latency, a little less than the 0.55\,ms of one expert copy.

\section{Cache-Conditioned Routing and Transfer Timing}
\label{sec:semantics}

Combining \cref{eq:cp} with prefetching is natural: prefetched experts become resident, receive the bonus, and are then likely to be selected. But if residency is the \emph{physical} state of the GPU cache, the routing at token $t$ depends on which copies have been issued and completed by then, which depends on bandwidth, contention, and scheduler parameters.

\Cref{tab:determinism} makes this concrete. We run \cp routing ($\lambda=0.2$, 50\% residency) with the A eviction priority of \cref{sec:design} and cross-layer prefetch, where routing sees the experts whose copies the scheduler has issued, and vary only the number of prefetch copies the scheduler keeps in flight (1, 2 or 4). This parameter changes nothing about the model, yet perplexity moves from \detPhysOneA\ to \detPhysFourA, because a different set of experts is resident when the router runs. The same happens at 25\% residency with $\lambda=0.25$ (\detPhysOneB\ to \detPhysFourB). Even with all parameters fixed, repeated runs on the same GPU differ by up to \detPhysRngA\ (50\%) and \detPhysRngB\ (25\%) in perplexity. A GPU with different PCIe bandwidth, or the same GPU under different load, can be expected to produce yet another model. A harness that does not reproduce the engine's timing then evaluates a different policy from the one deployed. Such a system can still be evaluated, by benchmarking inside the deployed engine or by fixing the order of copies and computation, but the result then holds for one hardware and load configuration.

\begin{table}[t]
\caption{Perplexity (12 windows) of \cp routing with A eviction and cross-layer prefetch of 8 when only the number of prefetch copies in flight changes; mean of \nRep\ runs, with the range across runs in parentheses. Physical cache: routing sees the experts whose copies have been issued (the scheduler also prefetches by token prior). Logical cache: routing sees the logical cache of \psr (unbounded insertion). With routing on the physical cache, outputs depend on the scheduler parameter and vary between runs; with the logical cache they do not.}
\label{tab:determinism}
\centering
\small
\setlength{\tabcolsep}{3pt}
\begin{tabular}{@{}lrrr@{}}
\toprule
prefetch copies in flight & 1 & 2 & 4 \\
\midrule
\multicolumn{4}{@{}l}{\emph{$C=32$, $\lambda=0.2$, cross-layer prefetch of 8}} \\
physical cache & 8.755\,{\scriptsize(0.041)} & 8.674\,{\scriptsize(0.054)} & 8.627\,{\scriptsize(0.048)} \\
logical cache & 8.509\,{\scriptsize(0.000)} & 8.509\,{\scriptsize(0.000)} & 8.509\,{\scriptsize(0.000)} \\
\midrule
\multicolumn{4}{@{}l}{\emph{$C=16$, $\lambda=0.25$, cross-layer prefetch of 8}} \\
physical cache & 9.097\,{\scriptsize(0.013)} & 9.015\,{\scriptsize(0.013)} & 8.864\,{\scriptsize(0.027)} \\
logical cache & 8.587\,{\scriptsize(0.000)} & 8.587\,{\scriptsize(0.000)} & 8.587\,{\scriptsize(0.000)} \\
\bottomrule
\end{tabular}

\end{table}

We instead separate the cache that routing sees from the cache that holds bytes. The routing policy maintains a \emph{logical} cache whose evolution depends only on the token history and the policy's parameters. The engine keeps physical residency consistent with it by issuing copies, and waits for a copy only when an expert it must execute has not arrived. Timing then changes how long a token takes, never what it computes. The evaluation harness simulates the same logical cache, so the harness and the engine implement the same routing policy and differ only in floating-point numerics (\cref{sec:impl}); \cref{sec:engine-quality} checks how far benchmark comparisons carry over by generating three benchmarks with the engine itself. The bottom rows of \cref{tab:determinism} confirm that the logical variant is invariant to the scheduler parameter. Determinism here is relative to fixed numerics: the logical cache removes the dependence on timing, not the dependence on kernels, batch composition or hardware arithmetic, which affect any implementation, including exact routing.

\section{Prefetch-Steered Routing}
\label{sec:design}

\subsection{Per-layer step}

\psr keeps, per layer, a logical residency set $R_l$ of size at most $C$, an eviction priority, and the statistics that define it. It is reset to a fixed warm set at the start of every sequence (the $C$ most frequently selected experts of the layer on held-out text), so each request is processed independently. \Cref{alg:psr} lists the step for token $t$ at layer $l$.

\begin{algorithm}[t]
\caption{\psr logical-cache step for token $t$. Part 1 runs at layer $l{-}1$, as soon as that layer's MoE computation has been launched; part 2 runs at layer $l$, once its router logits are available. Layer 1 has no pre-gate and skips part 1.}
\label{alg:psr}
\begin{algorithmic}[1]
\STATE $B \leftarrow \emptyset$ \hfill\COMMENT{stays empty at layer 1}
\STATE \textbf{Part 1, at layer $l{-}1$: insertion for layer $l$}
\STATE \textbf{input:} pre-gate ranking $g$ (router $l$ on layer $l{-}1$'s MoE input), priority $p$ of layer $l$ (fixed for token $t$)
\STATE $\textit{new} \leftarrow 0$
\FOR{$e$ in $g[1{:}8]$ (in rank order)}
  \IF{$e \in R_l$}
    \STATE $B \leftarrow B \cup \{e\}$
  \ELSIF{$\textit{new} < n$}
    \STATE if $|R_l| = C$, evict $\arg\min_{j\in R_l\setminus B} p_j$
    \STATE $R_l \leftarrow R_l \cup \{e\}$; $B \leftarrow B \cup \{e\}$; $\textit{new} \mathrel{+}= 1$; queue a copy of $e$
  \ENDIF
\ENDFOR
\STATE \textbf{Part 2, at layer $l$: selection}
\STATE \textbf{input:} router logits $z$
\STATE $S \leftarrow \mathrm{TopK}\big(z + \lambda\bar\Delta_l\,\mathbb{1}[\cdot\in R_l]\big)$ \hfill\COMMENT{\cref{eq:cp} on $R_l$}
\FOR{$e \in S \setminus R_l$ (in index order)}
  \STATE evict $\arg\min_{j\in R_l\setminus (S\cup B)} p_j$, or $\arg\min_{j\in R_l\setminus S} p_j$ if that set is empty
  \STATE $R_l\leftarrow R_l\cup\{e\}$; issue a demand copy of $e$
\ENDFOR
\STATE execute $S$ with weights $\mathrm{softmax}(z)_S$; wait only for copies of $S$ not yet complete
\STATE update the statistics of layer $l$ with the router's own top-$K$ of $z$
\end{algorithmic}
\end{algorithm}

\textbf{Bounded prefetch steering (part 1).} The candidates are the first 8 experts of the pre-gate ranking, i.e.\ layer $l$'s router applied one layer early. They are visited in rank order, and every candidate visited so far, resident or newly inserted, is protected from eviction for the rest of the step. A resident candidate further down the ranking is not yet protected, so an insertion can evict it if it has the lowest priority; with the A priority this happens in \insLaterAA\% of insertions at 50\% residency and \insLaterAB\% at 25\% (\insLaterLruA\% and \insLaterLruB\% with LRU), because the priority already favors experts the router has been choosing. At most $n$ new candidates are inserted. Each insertion makes the expert eligible for the routing bonus in part 2, which is how prefetch \emph{steers} routing: the bias now pulls the selection toward experts that are already in flight and that the router itself ranks highly. Without the budget, all non-resident candidates would be inserted; this gives the closest match to exact routing but commits more copies than the link can overlap (\cref{sec:ablations}). One layer of computation is shorter than one copy, so even with $n=1$ most prefetches are still in flight when their layer starts; they shorten the wait rather than remove it (\cref{sec:eval-why}).

\textbf{Selection (part 2)} is \cref{eq:cp} on the logical set. Ties are broken toward the router's own ranking. Execution weights come from the unmodified softmax, as in \cp. Demand insertions spare the experts inserted or visited in part 1 while any other victim outside $S$ exists. All remaining ties in priority go to the lowest expert index, so every decision is a function of the token history.

\textbf{Eviction priority.} We use the following score of an expert as its priority, and call it the \emph{A priority}:
\[
  p_e = \bar s^{\text{fast}}_e + \bar s^{\text{slow}}_e + \pi_l(x_t)_e ,
\]
where $\bar s^{\text{fast}}$ and $\bar s^{\text{slow}}$ are exponential moving averages (decay 0.5 and 0.98) of the indicator that the router's own top-$K$ contains $e$, and $\pi_l(x)$ is a token prior, the frequency with which layer $l$ routes input token $x$ to each expert on held-out text. The averages track the router's preference rather than the executed set, so the bias does not reinforce itself through eviction. The priority is computed once per step, before any insertion; ties go to the lowest index.

\subsection{Physical scheduling}

The engine mirrors \cref{alg:psr} and issues a copy for every logical insertion. Demand copies are issued immediately. Prefetch copies are queued and issued at layer boundaries with at most two in flight, so that a later demand copy waits behind at most two prefetches. A queued prefetch whose slot has been reassigned before it was issued (its expert was evicted again) is dropped. All copies go through one ordered copy stream, so a later copy into a slot completes after an earlier one, and each slot records the event of its latest copy. The computation of a layer waits only on the events of the slots it executes. Because all decisions are made on logical state, the scheduler's choices, including how many copies are in flight, affect latency alone. At the start of a request the logical cache is reset to the warm set; the engine copies the warm experts that are not physically in place before processing the prompt, so this cost is part of prompt processing, which our decode latencies exclude.

\textbf{Speculative execution as a special case.} If line 12 is replaced by $S = g[1{:}K]$, the layer executes the predicted experts instead of the router's choice, which is a variant of the speculative-execution strategy of \citet{madan2026speculating}. It differs from theirs in two ways: the prediction uses the next layer's router on the current hidden state, without their learned default vector, and the predicted experts are weighted by the true router softmax rather than by predicted weights. We implement it in the same engine and logical cache as a baseline.

\section{Implementation}
\label{sec:impl}

\textbf{Engine.} The decode engine is written in PyTorch for OLMoE. Per layer it replays two CUDA graphs: attention, router and pre-gate (whose outputs are copied to pinned host memory), then the MoE computation with the fused MoE kernel of vLLM~\citep{kwon2023vllm} over a slot-indexed expert cache. Between the two graphs the host runs the routing policy (the cache model of \cref{alg:psr} compiled with numba, the biased top-$K$ selection in numpy at about 20\,\textmu s per layer, which \cp and \psr pay alike) and issues copies on a dedicated stream with events. All $L\times E$ experts are in pinned host memory. The KV cache is static. Ports to the DeepSeek-V3 and Qwen3.6 architectures serve the second and third models (\cref{sec:moonlight,sec:qwen}).

\textbf{Harness.} Benchmarks run in a HuggingFace implementation of OLMoE whose MoE blocks call the same logical cache model, with state carried across decoding steps and reset per sequence. With LRU eviction and no prefetching, the logical model reproduces \cp routing decision for decision (unit-tested). Engine and harness agree on perplexity: 8.684 vs 8.683 for \psr $\lambda=0.2$, $n=1$ at 50\% residency, and 9.009 vs 9.069 for $\lambda=0.25$, $n=1$ at 25\%. Selected expert sets agree on 75--92\% of token-layer pairs. The rest are bf16 near-ties: even exact routing differs between the two implementations on 4.75\% of token-layer pairs because the kernels accumulate in different orders, and a cache-conditioned router propagates a flipped near-tie through its state. Such differences are numerical, not timing-dependent: with the logical cache, the engine's perplexity is identical across repeated runs and across scheduler settings (\cref{tab:determinism}).

\section{Evaluation}
\label{sec:eval}

\subsection{Setup}
\label{sec:setup}

\textbf{Model and hardware.} OLMoE-1B-7B-0924-Instruct~\citep{muennighoff2025olmoe} has 16 MoE layers with 64 experts each and top-8 routing. One expert is 12.6\,MB in bf16; all 1024 experts take 12.9\,GB. We use one NVIDIA RTX~4090 (24\,GB) on PCIe~4.0~$\times$16, where one expert copy takes 0.55\,ms (22.9\,GB/s measured). The engine process is pinned to 16 CPU cores. Experts stay in bf16; quantization is orthogonal and not studied.

\textbf{Residency.} We cap the expert cache at $C=32$ (50\%) or $C=16$ (25\%) slots per layer. Because the full model would fit on this GPU, this emulates a smaller GPU or a larger model with the same per-layer ratio.

\textbf{Latency.} We decode 12 windows of up to 1024 tokens (4 each from WikiText, C4, and text sampled from the model itself) token by token with teacher forcing, which executes the batch-1 decode path for every position and lets every configuration process the same tokens. We report mean and 99th-percentile per-token latency over about 9.8k tokens, and expert copies per token, split into demand copies that stall the layer and prefetch copies that overlap computation. Timings are the mean of \nRep\ repetitions; repetition $r$ of every configuration ran in the same timing round. Configurations added in this revision (the $n=1$ ablation and \cp with top-$J$ protection or a running bonus scale) were timed in a second session with the same protocol and also three rounds; latency ratios and tests pair repetitions only within a session. We report latency ratios with 95\% intervals and $p$-values from a paired $t$-test on log latency over repetitions, which treats the repetition, not the token, as the unit~\citep{kalibera2013rigorous}. Timing processes ran concurrently on separate GPUs, each with its CPU cores and pinned host memory on its GPU's NUMA node. With all experts resident, decoding takes \msFloor\ ms/token.

\textbf{Quality.} We use lm-evaluation-harness~\citep{gao2023lmeval} with greedy decoding at batch size 1 (on Moonlight and Qwen3.6 in batches, with one logical cache per row; \cref{app:protocol}), following each task's default protocol: MMLU~\citep{hendrycks2021mmlu} (0-shot, 14{,}042 questions), ARC-Challenge~\citep{clark2018arc} and HellaSwag~\citep{zellers2019hellaswag} (0-shot, accuracy), GSM8K~\citep{cobbe2021gsm8k} (5-shot, strict match), BBH~\citep{suzgun2023bbh} (3-shot chain-of-thought, 100 questions from each of 27 subtasks), IFEval~\citep{zhou2023ifeval} (chat template, strict prompt- and instruction-level accuracy), HumanEval~\citep{chen2021humaneval} and MBPP~\citep{austin2021mbpp} (pass@1). The last five are generative and exercise the routing policy over hundreds of decoding steps; MMLU, ARC and HellaSwag are scored from a single forward pass and stress the policy over the prompt. For each task we compare a configuration with a reference on the same questions and report the mean paired difference, its standard error, and an exact McNemar test~\citep{mcnemar1947note} on discordant questions~\citep{dror2018hitchhiker}. IFEval's instruction-level accuracy counts several instructions per prompt; for it the prompt is the unit, with a cluster-robust standard error and a sign-flip permutation test on per-prompt sums. Each table is one family of tests: \textbf{bold} marks differences significant after Holm's correction~\citep{holm1979simple} over all tests of the table, and $^\dagger$ marks $p<0.05$ before correction only, which we treat as exploratory. Comparisons reported only in the text are corrected over the tasks of that comparison. As a one-number summary we average the difference over the eight tasks (IFEval at prompt level) with a 95\% interval that treats tasks as independent. A difference that is not significant does not establish equivalence~\citep{lakens2017equivalence}. Where we call two configurations similar in quality, we give this interval and its lower end, the largest mean loss the data do not exclude, and leave the acceptable loss to the reader. Greedy generation is sensitive to any routing change (16--25\% of BBH items flip relative to exact routing under each configuration in \cref{tab:bench}), so the paired standard error is 0.8--1.0 points on BBH, 1.2--1.3 on GSM8K, 1.5--1.8 on IFEval (prompt level), 1.1--1.7 on MBPP and 1.8--3.0 on HumanEval.

\textbf{Diagnostic.} We also report the excess perplexity on 44 windows of text sampled from the exact model at temperature~1 (\emph{selfgen}). Because these samples come from the exact model, the excess negative log-likelihood estimates the KL divergence from the exact model. We use it to trace frontiers cheaply, not to accept configurations.

\textbf{Baselines.} All baselines run in the same engine. \emph{Exact routing} with LRU, with the A priority of \psr, and with the best timing-dependent cache and prefetch configuration we found (A priority plus throttled cross-layer prefetch; exact routing makes timing harmless for outputs). \emph{\cp}~\citep{skliar2025cache} with LRU, as in the original, for $\lambda\in\{0.1,\dots,0.3\}$. The original also keeps the router's top-$J$ experts ($J=2$ for fine-grained models) and tracks $\bar\Delta_l$ with a running average. Our main comparisons use $J=0$ and a fixed $\bar\Delta_l$ for \cp and \psr alike, which isolates the bonus; \cref{sec:ablations} also evaluates \cp as published, with $J\in\{1,2\}$ and a running average of the logit range over each request. \emph{Speculative execution}~\citep{madan2026speculating}, implemented as described in \cref{sec:design} with the same logical cache and prefetching as \psr; the shared cache helps it relative to the original, which keeps no experts resident across tokens, while our simpler predictor and the true router weights differ from theirs (\cref{sec:design}).

\textbf{Configuration selection.} We chose $\lambda$, $n$ and the pre-gate window on the selfgen diagnostic and engine latency (\cref{fig:frontier}); no evaluation data entered the token prior, the warm set or $\bar\Delta_l$ (\cref{app:configs}). We then ran the full benchmark suite on 7 \cp and 6 \psr configurations of OLMoE and on speculative execution. The operating points of \cref{tab:bench,tab:pairs} were fixed after seeing these results: \cp $\lambda=0.2$ as the largest $\lambda$ without a significant loss on a generative task, and the \psr points as those closest to it in latency or in quality. The three \cp configurations not shown ($\lambda=0.15$ and $0.25$ at 50\% residency, $0.25$ at 25\%) lie between the ones shown. The tests are paired and corrected for multiplicity, but the operating points were not pre-registered.

\subsection{Latency}
\label{sec:eval-latency}

\begin{table*}[t]
\caption{Engine latency on OLMoE-1B-7B, RTX 4090. sg\% = selfgen excess perplexity (KL diagnostic, lower is closer to exact routing). ms = mean per-token latency, p99 = 99th percentile, copies = expert copies per token; in parentheses, demand copies, issued only when the layer needs the expert. Prefetch copies that are still in flight when needed also delay the layer (\cref{fig:xfer}). $^\dagger$Routing or caching depends on transfer timing; for exact routing this affects latency only. \psr rows use A eviction (except the row marked LRU) and cross-layer prefetch with a budget of $n$ new insertions per layer and token.}
\label{tab:latency}
\centering
\small
\begin{tabular}{@{}lrrrrrrrr@{}}
\toprule
 & \multicolumn{4}{c}{$C=32$ (50\%)} & \multicolumn{4}{c}{$C=16$ (25\%)} \\
\cmidrule(lr){2-5}\cmidrule(lr){6-9}
config & sg\% & ms & p99 & copies (stall) & sg\% & ms & p99 & copies (stall) \\
\midrule
all experts resident (C=64) & 0.0 & 6.8 & 7 & 0.0 (0.0) & \multicolumn{4}{c}{(same)} \\
\midrule
exact routing, LRU & 0.0 & 24.7 & 52 & 30.5 (30.5) & 0.0 & 42.5 & 69 & 61.4 (61.4) \\
exact routing, A eviction & 0.0 & 22.9 & 48 & 27.5 (27.5) & 0.0 & 41.4 & 68 & 59.2 (59.2) \\
exact routing, best cache + prefetch$^\dagger$ & 0.0 & 20.5 & 43 & 30.7 (16.9) & 0.0 & 37.8 & 63 & 62.5 (42.6) \\
exact routing, logical prefetch ($n{=}8$) & 0.0 & 20.9 & 44 & 31.6 (16.7) & 0.0 & 39.0 & 64 & 65.2 (43.8) \\
\midrule
Cache-Prior $\lambda$0.2, LRU & 6.3 & 14.3 & 30 & 10.5 (10.5) & 7.9 & 27.5 & 51 & 33.1 (33.1) \\
Cache-Prior $\lambda$0.3, LRU & 18.2 & 11.1 & 21 & 5.0 (5.0) & 22.3 & 20.7 & 39 & 21.4 (21.4) \\
Cache-Prior, A eviction ($\lambda$0.2 / 0.25) & 6.1 & 13.8 & 30 & 9.7 (9.7) & 12.5 & 23.9 & 46 & 26.3 (26.3) \\
Cache-Prior $\lambda$0.2 + timing-dep.\ prefetch$^\dagger$ & -- & 13.9 & 27 & 18.3 (4.5) & -- & 24.0 & 44 & 38.3 (18.4) \\
\midrule
speculative execution & 5.4 & 20.0 & 43 & 29.1 (13.3) & 5.4 & 38.1 & 64 & 63.5 (41.9) \\
\midrule
PSR $\lambda$0.2, unbounded ($n{=}8$) & 1.4 & 17.0 & 37 & 24.8 (9.2) & 2.5 & 33.6 & 59 & 55.7 (34.2) \\
PSR $\lambda$0.2, LRU eviction, $n{=}8$ & 1.7 & 18.7 & 43 & 27.9 (11.8) &  & & &  \\
PSR $\lambda$0.2, $n{=}2$ & 2.2 & 14.5 & 27 & 20.8 (5.2) & 4.9 & 24.7 & 45 & 41.0 (19.4) \\
PSR $\lambda$0.2, $n{=}1$ & 3.3 & 12.0 & 25 & 16.3 (4.6) & 6.3 & 21.8 & 43 & 35.6 (20.9) \\
PSR $\lambda$0.25, $n{=}1$ & 4.1 & 11.3 & 22 & 14.9 (3.3) & 8.8 & 19.2 & 39 & 31.1 (16.4) \\
PSR $\lambda$0.3, $n{=}1$ & 5.4 & 10.7 & 19 & 13.8 (2.2) & 12.7 & 16.9 & 34 & 27.0 (12.3) \\
\bottomrule
\end{tabular}

\end{table*}

\begin{figure*}[t]
\centering
\includegraphics[width=\textwidth]{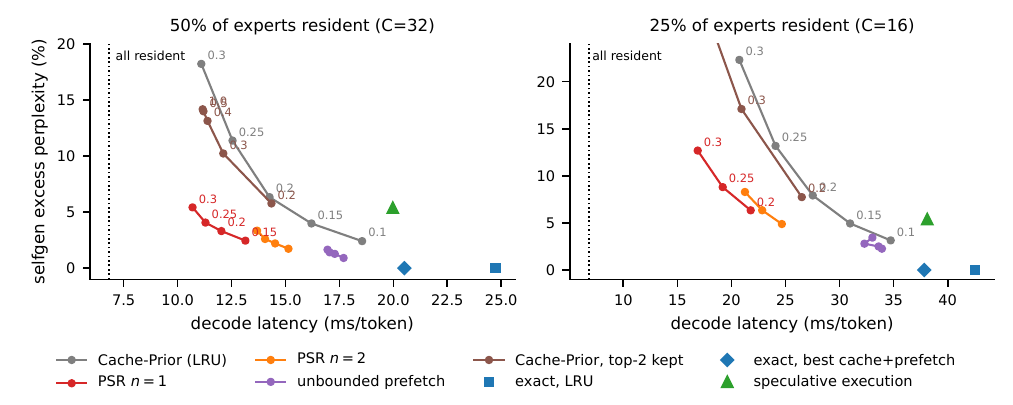}
\caption{Latency vs.\ divergence from exact routing (selfgen excess, a KL diagnostic) for $\lambda\in\{0.1,0.15,0.2,0.25,0.3\}$ (labels on \cp and $n{=}1$ curves). \psr with a budget of one insertion per layer and token is the fastest configuration at every divergence level its $\lambda$ sweep covers; unbounded prefetch is closest to exact but slow. Quality claims rest on the benchmarks of \cref{fig:bench}.}
\label{fig:frontier}
\end{figure*}

\Cref{tab:latency} and \cref{fig:frontier} summarize the engine measurements. At 50\% residency, exact routing with LRU takes \msExactLruA\ ms/token. A better eviction priority and throttled cross-layer prefetch bring exact routing to \msExactBestA\ ms, but the link stays saturated: about 30 copies per token, of which more than half still stall. \cp at $\lambda=0.2$, the largest $\lambda$ we ran on the full suite at 50\% residency without a significant loss on any generative task, reduces copies to \stallCPtwoA\ per token and takes \msCPtwoA\ ms. \psr ($\lambda=0.3$, $n=1$) takes \msFastA\ ms with a p99 of \tailFastA\ ms (\cp: \tailCPtwoA\ ms; exact LRU: \tailExactLruA\ ms): exact routing with LRU takes \ratExactLruA$\times$ as long (95\% interval \ratExactLruAlo--\ratExactLruAhi, $p=\ratExactLruAp$), the best exact configuration \ratExactBestA$\times$ (\ratExactBestAlo--\ratExactBestAhi), and \cp $\lambda=0.2$ \ratCPA$\times$ (\ratCPAlo--\ratCPAhi). At 25\% residency, the ordering is the same: \msExactLruB\ ms for exact LRU, \msCPtwoB\ ms for \cp $\lambda=0.2$, and \msFastB\ ms for \psr ($\lambda=0.25$, $n=1$).

\begin{figure}[t]
\centering
\includegraphics[width=\columnwidth]{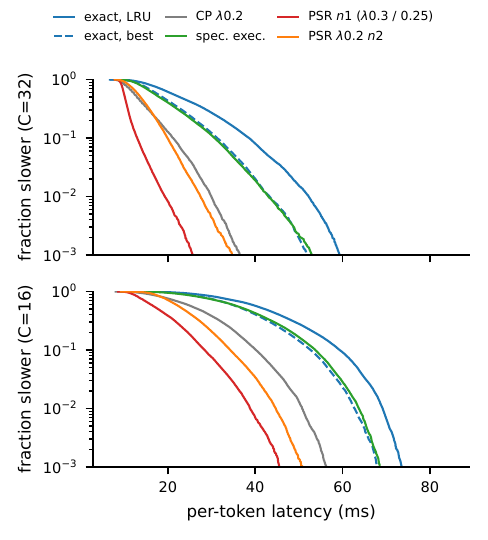}
\caption{Distribution of per-token decode latency at 50\% (top) and 25\% (bottom) residency: fraction of tokens slower than $x$ (log scale), over all tokens of the 3 timing repetitions.}
\label{fig:cdf}
\end{figure}

\Cref{fig:cdf} shows the full per-token distributions. From the 10th percentile up, \psr is faster than every baseline shown, and its gap to \cp $\lambda=0.2$ is larger in the tail than at the median.

Across the $\lambda$ sweep, \psr with $n=1$ lies below and to the left of \cp at every point (\cref{fig:frontier}). At the same $\lambda$ it is both faster and closer to exact routing: at $\lambda=0.3$ and 50\% residency, \cp has a selfgen excess of \sgCPthreeA\% and \psr \sgFastA\%. The routing bonus is the same; what differs is which experts hold it.

\subsection{Quality}
\label{sec:eval-quality}

\begin{table*}[t]
\caption{Benchmark scores relative to exact routing (points; paired over identical questions; \textbf{bold}: significant after Holm's correction over the table's \nTestsBench\ tests; $^\dagger$: $p<0.05$ before correction only). The first row gives absolute scores of exact routing. ms is engine latency at the configuration's residency. IFE-P/IFE-I = IFEval strict prompt/instruction accuracy. Mean = average over the eight tasks with IFE-P. Speculative execution's outputs do not depend on residency; its two rows share one benchmark run.}
\label{tab:bench}
\centering
\small
\setlength{\tabcolsep}{4pt}
\begin{tabular}{@{}llrrrrrrrrrrr@{}}
\toprule
$C$ & config & ms & MMLU & ARC-C & HellaS. & GSM8K & BBH & IFE-P & IFE-I & HumanE. & MBPP & mean \\
\midrule
-- & exact routing & 24.7/42.5 & 52.4 & 49.0 & 60.6 & 33.2 & 36.0 & 41.2 & 52.3 & 32.9 & 28.8 & -- \\
\midrule
32 & CP $\lambda$0.2 & 14.3 & $-$0.3 & $-$0.5 & \textbf{$-$0.9} & +0.3 & $-$0.6 & $-$2.2 & +0.5 & +3.0 & +0.2 & $-$0.13 \\
32 & CP $\lambda$0.3 & 11.1 & \textbf{$-$1.1} & $-$2.3$^\dagger$ & \textbf{$-$2.4} & +0.1 & $-$2.9$^\dagger$ & $-$2.4 & $-$1.8 & $-$1.8 & $-$0.6 & $-$1.69 \\
32 & PSR $\lambda$0.2 $n$1 & 12.0 & $-$0.7$^\dagger$ & +0.1 & $-$0.2 & +1.3 & $-$0.2 & $-$0.7 & $-$0.4 & +0.6 & $-$0.4 & $-$0.04 \\
32 & PSR $\lambda$0.3 $n$1 & 10.7 & $-$0.8$^\dagger$ & $-$0.5 & $-$0.4$^\dagger$ & +0.2 & $-$0.9 & $-$2.4 & $-$2.5 & +5.5$^\dagger$ & $-$0.8 & $-$0.03 \\
32 & spec. exec. & 20.0 & \textbf{$-$1.3} & +1.5 & \textbf{$-$0.9} & $-$0.7 & $-$1.9$^\dagger$ & $-$2.0 & $-$0.2 & +1.2 & +1.0 & $-$0.39 \\
\midrule
16 & CP $\lambda$0.2 & 27.5 & $-$0.8$^\dagger$ & $-$0.3 & \textbf{$-$0.8} & $-$1.9 & $-$2.3$^\dagger$ & +0.4 & +0.7 & $-$0.6 & $-$0.6 & $-$0.88 \\
16 & CP $\lambda$0.3 & 20.7 & \textbf{$-$2.0} & $-$0.7 & \textbf{$-$2.7} & $-$3.7$^\dagger$ & \textbf{$-$4.9} & $-$3.3 & $-$1.9 & $-$3.7 & +0.2 & $-$2.60 \\
16 & PSR $\lambda$0.2 $n$2 & 24.7 & $-$0.6$^\dagger$ & +1.0 & $-$0.4$^\dagger$ & +0.3 & $-$1.7$^\dagger$ & $-$0.9 & +0.4 & +0.6 & $-$1.8 & $-$0.43 \\
16 & PSR $\lambda$0.25 $n$1 & 19.2 & $-$0.7$^\dagger$ & $-$0.4 & \textbf{$-$0.8} & $-$0.3 & $-$2.6$^\dagger$ & +0.2 & +1.9 & +1.8 & 0.0 & $-$0.36 \\
16 & spec. exec. & 38.1 & \textbf{$-$1.3} & +1.5 & \textbf{$-$0.9} & $-$0.7 & $-$1.9$^\dagger$ & $-$2.0 & $-$0.2 & +1.2 & +1.0 & $-$0.39 \\
\bottomrule
\end{tabular}

\end{table*}

\begin{table*}[t]
\caption{Direct paired comparisons of \psr with \cp (and speculative execution) on identical questions (points; \textbf{bold}: significant after Holm's correction over the table's \nTestsPairs\ tests; $^\dagger$: $p<0.05$ before correction only). Mean $\pm$ 95\% interval over the eight tasks. Iso-latency pairs run at about the same ms/token; the other pairs compare \psr with the \cp operating point of \cref{tab:bench} ($\lambda=0.2$), which is slower. A mean difference that is not significant does not show equal quality; its 95\% lower bound limits how much worse \psr can be on average. Speculative execution does not depend on residency, so one benchmark run serves both $C$.}
\label{tab:pairs}
\centering
\small
\setlength{\tabcolsep}{2.7pt}
\begin{tabular}{@{}lllrrrrrrrrrr@{}}
\toprule
$C$ & PSR (ms) & vs. (ms) & MMLU & ARC-C & HellaS. & GSM8K & BBH & IFE-P & IFE-I & HumanE. & MBPP & mean \\
\midrule
\multicolumn{13}{@{}l}{\emph{iso-latency}} \\
32 & PSR $\lambda$0.3 $n$1 (10.7) & CP $\lambda$0.3 (11.1) & +0.3 & +1.8 & \textbf{+2.0} & +0.1 & +2.0$^\dagger$ & 0.0 & $-$0.7 & +7.3$^\dagger$ & $-$0.2 & +1.7 $\pm$1.0 \\
16 & PSR $\lambda$0.25 $n$1 (19.2) & CP $\lambda$0.3 (20.7) & \textbf{+1.2} & +0.3 & \textbf{+1.9} & +3.4$^\dagger$ & +2.3$^\dagger$ & +3.5 & +3.8$^\dagger$ & +5.5 & $-$0.2 & +2.2 $\pm$1.0 \\
\midrule
\multicolumn{13}{@{}l}{\emph{vs. CP operating point}} \\
32 & PSR $\lambda$0.3 $n$1 (10.7) & CP $\lambda$0.2 (14.3) & $-$0.5 & 0.0 & +0.4$^\dagger$ & $-$0.2 & $-$0.3 & $-$0.2 & $-$3.0 & +2.4 & $-$1.0 & +0.1 $\pm$0.9 \\
32 & PSR $\lambda$0.2 $n$1 (12.0) & CP $\lambda$0.2 (14.3) & $-$0.4 & +0.6 & \textbf{+0.6} & +1.0 & +0.4 & +1.5 & $-$0.8 & $-$2.4 & $-$0.6 & +0.1 $\pm$0.8 \\
16 & PSR $\lambda$0.25 $n$1 (19.2) & CP $\lambda$0.2 (27.5) & +0.1 & $-$0.2 & +0.1 & +1.6 & $-$0.3 & $-$0.2 & +1.2 & +2.4 & +0.6 & +0.5 $\pm$0.8 \\
16 & PSR $\lambda$0.2 $n$2 (24.7) & CP $\lambda$0.2 (27.5) & +0.3 & +1.3 & +0.4$^\dagger$ & +2.2 & +0.6 & $-$1.3 & $-$0.4 & +1.2 & $-$1.2 & +0.4 $\pm$0.9 \\
\midrule
\multicolumn{13}{@{}l}{\emph{vs. spec.}} \\
32 & PSR $\lambda$0.3 $n$1 (10.7) & spec. exec. (20.0) & +0.5 & $-$2.0 & +0.5$^\dagger$ & +0.8 & +1.0 & $-$0.4 & $-$2.3 & +4.3 & $-$1.8 & +0.4 $\pm$0.9 \\
32 & PSR $\lambda$0.2 $n$1 (12.0) & spec. exec. (20.0) & +0.6$^\dagger$ & $-$1.4 & +0.7$^\dagger$ & +2.0 & +1.7$^\dagger$ & +1.3 & $-$0.1 & $-$0.6 & $-$1.4 & +0.4 $\pm$0.9 \\
16 & PSR $\lambda$0.25 $n$1 (19.2) & spec. exec. (38.1) & +0.5 & $-$1.9 & +0.1 & +0.4 & $-$0.7 & +2.2 & +2.2 & +0.6 & $-$1.0 & 0.0 $\pm$1.0 \\
\bottomrule
\end{tabular}

\end{table*}

\begin{figure}[t]
\centering
\includegraphics[width=\columnwidth]{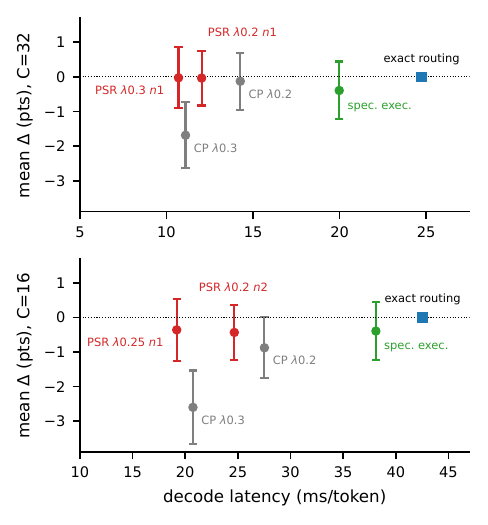}
\caption{Mean benchmark difference to exact routing over eight tasks ($\pm$95\% interval) against decode latency, at 50\% (top) and 25\% (bottom) residency.}
\label{fig:bench}
\end{figure}

\Cref{tab:bench,tab:pairs} and \cref{fig:bench} report the benchmarks. Three comparisons matter.

\textbf{Against \cp at the same latency, \psr is better.} At 50\% residency, \psr ($\lambda=0.3$, $n=1$, \msFastA\ ms) and \cp ($\lambda=0.3$, \msCPthreeA\ ms) run at the same speed (latency ratio \ratIsoA, $p=\ratIsoAp$). \psr is ahead by \pairIsoLatA\ points on average (95\% interval $\pm$\pairIsoLatAci); after correction over the \nTestsPairs\ tests of \cref{tab:pairs} its gain on HellaSwag is significant, BBH and HumanEval are nominally so, and no task is worse. At 25\% residency the gap is larger, \pairIsoLatB\ $\pm$\pairIsoLatBci\ points, significant on MMLU and HellaSwag after correction and nominally on GSM8K, BBH and IFEval (instruction level). Excluding HumanEval, the noisiest task, the averages are \pairIsoLatAnohe\ and \pairIsoLatBnohe.

\textbf{Against \cp at similar quality, \psr is faster.} At 50\% residency, \psr with $\lambda=0.2$ (\msQualA\ ms) and \cp with $\lambda=0.2$ (\msCPtwoA\ ms) differ by \pairIsoQualAq\ $\pm$\pairIsoQualAqci\ points on average (lower end \pairIsoQualAqlo, or \pairIsoQualAqnohelo\ without HumanEval), and \cp takes \ratCPAq$\times$ as long (\ratCPAqlo--\ratCPAqhi). The faster \psr point (\msFastA\ ms, \redCPA\% lower latency than \cp $\lambda=0.2$) has the same mean difference, \pairIsoQualA\ $\pm$\pairIsoQualAci\ (lower end \pairIsoQualAlo; \pairIsoQualAnohelo\ without HumanEval), but larger point losses on IFEval at instruction level ($-3.0$) and MMLU ($-0.5$), both with $p\approx0.06$. At 25\% residency, \psr at \msFastB\ ms and \cp $\lambda=0.2$ at \msCPtwoB\ ms (\redCPB\% lower latency) differ by \pairIsoQualB\ $\pm$\pairIsoQualBci\ (lower end \pairIsoQualBlo), and no task differs even before correction.

\textbf{Against exact routing}, neither 50\% \psr configuration differs significantly on any single task after correction over the \nTestsBench\ tests of \cref{tab:bench}, whereas \cp loses significantly on HellaSwag at $\lambda=0.2$ and on MMLU and HellaSwag at $\lambda=0.3$. The averages need care. The fast 50\% configuration has a mean of \meanExFastA\ $\pm$\meanExFastAci\ over eight tasks, but this includes a HumanEval score $5.5$ points above exact routing; without HumanEval its mean is \meanExFastAnohe\ $\pm$\meanExFastAnoheci, a small loss spread over the other tasks. The slower 50\% point ($\lambda=0.2$, $n=1$, \msQualA\ ms) is at \meanExQualAnohe\ $\pm$\meanExQualAnoheci\ without HumanEval. At 25\% residency every cache-aware configuration we ran on BBH loses on it: 2.6 points for \psr at \msFastB\ ms ($p=0.002$), 1.7 for \psr $\lambda=0.2$, $n=2$ at \msQualB\ ms ($p=0.04$), and 2.3 for \cp $\lambda=0.2$ at \msCPtwoB\ ms ($p=0.01$). None of these survives the correction alone, but they agree in sign and size. At this residency no method is free on multi-step reasoning; \psr reaches a given BBH cost at lower latency than \cp, but does not remove it.

\textbf{HumanEval.} On HumanEval (164 problems) the fast 50\% \psr configuration scores $+5.5$ above exact routing ($p=0.035$, not significant after correction). We do not read this as an improvement, and we state every average above also without HumanEval.

\subsection{Benchmarks generated by the engine}
\label{sec:engine-quality}

The harness runs the engine's routing policy, but not its kernels: the engine uses CUDA graphs, FlashAttention and the fused MoE kernel, the harness HuggingFace's modules. We therefore generated GSM8K, IFEval and HumanEval with the engine itself (batch 1, greedy, the prompt fed token by token through the same policy) for exact routing, \cp $\lambda=0.2$ and the fast and quality \psr points ($\lambda=0.3$ and $0.2$, $n=1$) at 50\% residency, and compared them per question with the harness runs of the same configurations (\cref{tab:engbench}).

\textbf{Numerics alone change individual outputs.} Under exact routing, where the cache influences nothing, \ebDiscExactGsm\% of GSM8K questions are solved by one implementation and not by the other, with no net difference (\ebExactGsmeng\ vs.\ \ebExactGsmhar). Greedy decoding turns floating-point near-ties into different continuations. Neither implementation's per-question outputs predict the other's; what can carry over is the score of each configuration and the paired difference between configurations.

\textbf{No discrepancy in the paired differences is detected.} For each pair of configurations and each task we compare the per-question difference in the engine with that in the harness (sign-flip permutation test on the difference of differences; IFEval instructions clustered by prompt). None of the \ebNtests\ tests is significant after Holm's correction; one is nominally so (IFEval at instruction level, \psr vs.\ \cp, where the engine favours \psr by \ebNomDiff\ points more than the harness). The check has limited power: the 95\% intervals have half-widths of \ebCImin\ to \ebCImax\ points, so per-task discrepancies of that size would go undetected. The per-task differences of \cref{tab:pairs} are smaller than that and rest on the harness.

\begin{table}[t]
\caption{Engine-generated benchmarks against the harness at 50\% residency: paired difference between two configurations (points) in the harness and in the engine on the same questions, and the difference of the two with its 95\% interval ($^\dagger$: $p<0.05$ before Holm's correction over the table's \ebNtests\ tests; none survives it).}
\label{tab:engbench}
\centering
\small
\begin{tabular}{@{}lrrr@{}}
\toprule
task & harness & engine & engine $-$ harness \\
\midrule
\multicolumn{4}{@{}l}{\emph{PSR $\lambda$0.3 $n$1 vs.\ CP $\lambda$0.2}} \\
GSM8K & $-$0.2 & $-$0.8 & $-$0.7 $\pm$3.5 \\
IFEval-P & $-$0.2 & +1.7 & +1.8 $\pm$4.3 \\
IFEval-I & $-$3.0 & +0.7 & +3.7$^\dagger$ $\pm$3.5 \\
HumanEval & +2.4 & $-$1.8 & $-$4.3 $\pm$4.6 \\
\midrule
\multicolumn{4}{@{}l}{\emph{PSR $\lambda$0.2 $n$1 vs.\ CP $\lambda$0.2}} \\
GSM8K & +1.0 & $-$0.1 & $-$1.1 $\pm$3.5 \\
IFEval-P & +1.5 & +1.7 & +0.2 $\pm$4.3 \\
IFEval-I & $-$0.8 & +0.6 & +1.4 $\pm$3.6 \\
HumanEval & $-$2.4 & $-$2.4 & 0.0 $\pm$3.8 \\
\midrule
\multicolumn{4}{@{}l}{\emph{PSR $\lambda$0.3 $n$1 vs.\ exact, LRU}} \\
GSM8K & +0.2 & +0.8 & +0.7 $\pm$3.3 \\
IFEval-P & $-$2.4 & $-$0.2 & +2.2 $\pm$4.3 \\
IFEval-I & $-$2.5 & $-$0.1 & +2.4 $\pm$3.5 \\
HumanEval & +5.5$^\dagger$ & +4.3 & $-$1.2 $\pm$4.8 \\
\midrule
\multicolumn{4}{@{}l}{\emph{CP $\lambda$0.2 vs.\ exact, LRU}} \\
GSM8K & +0.3 & +1.7 & +1.4 $\pm$3.3 \\
IFEval-P & $-$2.2 & $-$1.8 & +0.4 $\pm$4.1 \\
IFEval-I & +0.5 & $-$0.8 & $-$1.3 $\pm$3.4 \\
HumanEval & +3.0 & +6.1$^\dagger$ & +3.0 $\pm$4.3 \\
\midrule
\multicolumn{4}{@{}l}{\emph{PSR $\lambda$0.2 $n$1 vs.\ exact, LRU}} \\
GSM8K & +1.3 & +1.6 & +0.3 $\pm$3.2 \\
IFEval-P & $-$0.7 & $-$0.2 & +0.6 $\pm$4.3 \\
IFEval-I & $-$0.4 & $-$0.2 & +0.1 $\pm$3.7 \\
HumanEval & +0.6 & +3.7 & +3.0 $\pm$4.3 \\
\bottomrule
\end{tabular}

\end{table}

\begin{table*}[t]
\caption{Latency in greedy free generation on \fgNprompts\ benchmark prompts (decode only; ms per generated token, its 99th percentile, and expert copies per token; means over \fgNrounds\ interleaved rounds), next to the teacher-forced latency (TF) of \cref{tab:latency}. Ratio rows: \psr over \cp per task, with the 95\% interval over rounds (paired on log latency), which measures run-to-run repeatability on this fixed prompt set; bold: significant after Holm's correction over the table's \fgNtests\ ratios. Over prompts: 95\% interval of the same ratio from resampling the task's prompts, and the number of prompts on which \psr is faster per token.}
\label{tab:freegen}
\centering
\small
\begin{tabular}{@{}lrrrrrrrrrr@{}}
\toprule
 & TF & \multicolumn{3}{c}{GSM8K} & \multicolumn{3}{c}{IFEval} & \multicolumn{3}{c}{HumanEval} \\
\cmidrule(lr){3-5}\cmidrule(lr){6-8}\cmidrule(lr){9-11}
config & ms & ms & p99 & cp. & ms & p99 & cp. & ms & p99 & cp. \\
\midrule
\multicolumn{11}{@{}l}{\emph{50\% residency}} \\
exact, LRU & 24.7 & 21.6 & 44.3 & 26.4 & 22.4 & 51.9 & 27.2 & 14.3 & 34.6 & 13.5 \\
CP $\lambda$0.2 & 14.3 & 10.8 & 23.4 & 5.2 & 12.1 & 29.5 & 7.5 & 8.1 & 14.6 & 1.4 \\
PSR $\lambda$0.3 $n$1 & 10.7 & 9.5 & 14.0 & 10.9 & 9.9 & 18.3 & 12.0 & 8.4 & 11.2 & 6.9 \\
PSR $\lambda$0.2 $n$1 & 12.0 & 10.2 & 19.2 & 12.3 & 10.7 & 22.8 & 13.0 & 8.6 & 13.4 & 7.1 \\
\quad PSR $\lambda$0.3 $n$1 / CP & & \multicolumn{3}{c}{\textbf{0.881} [0.840, 0.923]} & \multicolumn{3}{c}{\textbf{0.816} [0.750, 0.887]} & \multicolumn{3}{c}{\textbf{1.041} [1.031, 1.052]} \\
\qquad over prompts & & \multicolumn{3}{c}{[0.839, 0.929]; 20/20} & \multicolumn{3}{c}{[0.779, 0.863]; 15/16} & \multicolumn{3}{c}{[1.024, 1.060]; 7/41} \\
\quad PSR $\lambda$0.2 $n$1 / CP & & \multicolumn{3}{c}{\textbf{0.947} [0.900, 0.996]} & \multicolumn{3}{c}{\textbf{0.881} [0.811, 0.958]} & \multicolumn{3}{c}{\textbf{1.068} [1.061, 1.075]} \\
\qquad over prompts & & \multicolumn{3}{c}{[0.904, 0.998]; 13/20} & \multicolumn{3}{c}{[0.820, 0.939]; 14/16} & \multicolumn{3}{c}{[1.050, 1.083]; 4/41} \\
\midrule
\multicolumn{11}{@{}l}{\emph{25\% residency}} \\
CP $\lambda$0.2 & 27.5 & 21.0 & 44.9 & 22.4 & 23.1 & 51.3 & 26.5 & 13.7 & 35.0 & 10.6 \\
PSR $\lambda$0.25 $n$1 & 19.2 & 15.1 & 34.0 & 23.4 & 17.5 & 40.3 & 28.6 & 11.0 & 20.9 & 15.9 \\
\quad PSR $\lambda$0.25 $n$1 / CP & & \multicolumn{3}{c}{\textbf{0.718} [0.708, 0.728]} & \multicolumn{3}{c}{\textbf{0.759} [0.721, 0.799]} & \multicolumn{3}{c}{\textbf{0.806} [0.783, 0.829]} \\
\qquad over prompts & & \multicolumn{3}{c}{[0.683, 0.752]; 20/20} & \multicolumn{3}{c}{[0.710, 0.824]; 16/16} & \multicolumn{3}{c}{[0.755, 0.869]; 39/41} \\
\bottomrule
\end{tabular}

\end{table*}

\textbf{Free-generation latency depends on the workload.} The latencies of \cref{sec:eval-latency} are teacher-forced on WikiText, C4 and sampled model text. We also timed greedy generation in the engine on \fgNprompts\ benchmark prompts (every 66th GSM8K, 34th IFEval and 4th HumanEval problem: \fgNGsm, \fgNIfe\ and \fgNHe\ prompts), with the configurations of each residency level interleaved in one process over \fgNrounds\ rounds as in \cref{sec:eval-latency} (\cref{tab:freegen}). Each configuration decodes its own continuation; the logical cache makes it identical in every round. Every configuration is faster than under teacher forcing. At 25\% residency \psr keeps most of its advantage: it is \fgRedqGsm\%, \fgRedqIfe\% and \fgRedqHe\% faster than \cp on GSM8K, IFEval and HumanEval. At 50\% residency the gap narrows: the fast \psr point is \fgRedGsm\% (95\% interval \fgRedGsmci) faster on GSM8K and \fgRedIfe\% (\fgRedIfeci) on IFEval, but \fgSlowHe\% (\fgSlowHeci) slower on HumanEval. All \fgNtests\ ratios in \cref{tab:freegen} differ from one after Holm's correction. These intervals take the round as the unit and, since every round decodes the same continuations, measure run-to-run repeatability on this prompt set, not variation over prompts. Resampling prompts instead (rows ``over prompts'') also excludes one in all \fgNtests\ cases, though only approximately with \fgPrNmin--\fgPrNmax\ prompts per task; per-prompt sign tests survive Holm's correction except for \psr $\lambda=0.2$ on GSM8K (faster on \fgPrWinQGsm\ of \fgPrNQGsm). Each configuration is timed on its own continuations, which differ in content and length (on IFEval at 50\% residency, \fgPrLenPsrIfe\ tokens per prompt for the fast \psr point against \fgPrLenCPIfe\ for \cp). Single runs over the full benchmarks, not interleaved, give the same direction on every task and are less favourable to \psr by up to \fgFullDiff\ points.

The copy counts show why. Greedy benchmark outputs at 50\% residency reuse experts so heavily that \cp needs only \fgCPGsmcp, \fgCPIfecp\ and \fgCPHecp\ copies per token, while \psr's cross-layer insertions still cost \fgFastGsmcp, \fgFastIfecp\ and \fgFastHecp, mostly prefetches. On the prompt tokens of the same runs, which are not the model's own text, \psr is \fgRedPrGsm\%, \fgRedPrIfe\% and \fgRedPrHe\% faster than \cp. \psr's 99th-percentile latency is lower on every task and residency (at 50\%: \fgFastGsmpn\ vs.\ \fgCPGsmpn\ ms on GSM8K, \fgFastIfepn\ vs.\ \fgCPIfepn\ on IFEval, \fgFastHepn\ vs.\ \fgCPHepn\ on HumanEval). The latency advantage of \psr over \cp is therefore largest when routing has little locality (long or external text, long generations, low residency) and can vanish on short, highly repetitive outputs at moderate residency; against exact routing, \psr is \fgSpdExactRng$\times$ faster in free generation as well.

\subsection{Where the speedup comes from}
\label{sec:eval-why}

\begin{figure}[t]
\centering
\includegraphics[width=\columnwidth]{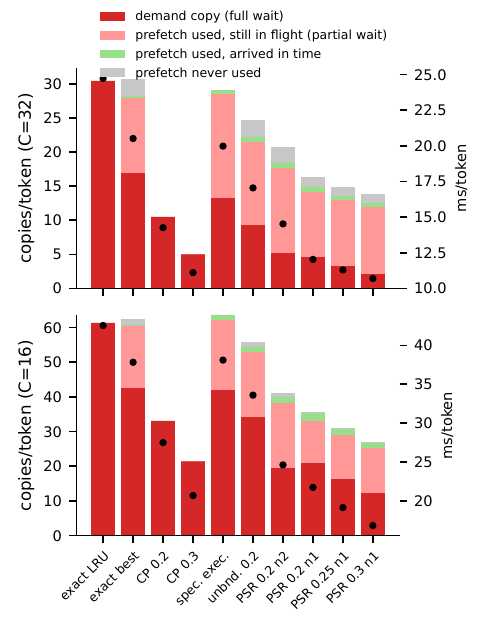}
\caption{Expert copies per token at 50\% (top) and 25\% (bottom) residency: demand copies, issued when the layer needs the expert (full wait); prefetch copies of executed experts that are still in flight when the layer is scheduled (partial wait) or that arrived in time; and prefetch copies whose expert was never executed. Dots: ms/token (right axis).}
\label{fig:xfer}
\end{figure}

\Cref{fig:xfer} decomposes the copies. \cp reduces the number of copies but leaves every remaining copy on the critical path: at $\lambda=0.2$ and 50\% residency, all \stallCPtwoA\ copies per token are demand copies, each a full 0.55\,ms wait. \psr makes \emph{more} copies (\xferFastA\ per token at $\lambda=0.3$, $n=1$), but only \stallFastA\ of them are demand copies. Of its \pfAllFastA\ prefetch copies per token, \pfUsedFastA\ bring an expert the layer executes; most of these (\lateFastA\ per token) were issued one layer earlier and are still in flight when the layer is scheduled, so the layer waits only for the remainder of the copy, and only \pfOnTimeFastA\ arrive in time. \pfUnusedFastA\ per token are never used. Per copy, the latency above the all-resident floor is \expCopyFastA\ ms for \psr, against \expCopyExactLruA\ ms for exact routing with LRU and \expCopyCPtwoA\ ms for \cp $\lambda=0.2$, where every copy is a demand copy that also stalls the host's scheduling of the layer. Speculative execution also prefetches one layer ahead, but because it executes the full predicted top-8, it needs as many copies as exact routing (\xferSpecA\ vs.\ \xferExactA\ per token): the link stays saturated, \stallSpecA\ copies per token end up as demand copies and \lateSpecA\ are late. \psr gets its speed from combining both mechanisms: the routing bonus cuts the number of copies the link must carry, and the prefetch budget starts the remaining ones early enough to overlap most of their latency.

\subsection{Ablations}
\label{sec:ablations}

\textbf{Prefetch budget.} Inserting all non-resident pre-gate candidates (unbounded, $n=8$) gives the configuration closest to exact routing (selfgen \sgUnbA\% at $\lambda=0.2$, 50\%), but commits about 25 copies per token, more than the link can overlap, and runs at \msUnbA\ ms, slower than \cp without prefetching. $n=2$ and $n=1$ trade divergence for speed along the frontier of \cref{fig:frontier}; $n=1$ is the fastest at every divergence level its $\lambda$ sweep covers.

\textbf{Steering and eviction at the operating point.} To separate the two mechanisms we timed all four combinations at $n=1$, $\lambda=0.2$ and 50\% residency in one session. \cp without prefetching runs at \ablCPLms\ ms with LRU and \ablCPAms\ ms with the A priority (selfgen \ablCPLsg\% and \ablCPAsg\%); adding steering gives \ablPsrLms\ ms with LRU and \ablPsrAms\ ms with the A priority (\ablPsrLsg\% and \ablPsrAsg\%). Most of the speedup comes from steering, which at the same $\lambda$ also roughly halves the divergence, because the inserted experts are ones the router ranks highly; the A priority adds a little speed and lowers the divergence further. With unbounded prefetch the A priority matters more: LRU raises latency from \msUnbA\ to \msUnbLruA\ ms and selfgen excess from \sgUnbA\ to \sgUnbLruA\%.

\textbf{\cp as published.} Keeping the router's top-2 experts bounds the divergence of \cp but also its savings, since the two experts must be loaded whatever the bonus. At 50\% residency, $J=2$ with $\lambda=0.2$ runs at \msJtwoTwoA\ ms with a selfgen excess of \sgJtwoTwoA\%, against \msFastA\ ms and \sgFastA\% for \psr ($\lambda=0.3$, $n=1$); raising $\lambda$ to 0.3 gives \msJtwoThreeA\ ms at \sgJtwoThreeA\%, and even $\lambda=1$ does not go below \msJtwoMaxA\ ms (\sgJtwoMaxA\%). At $\lambda=0.2$ the choice of $J$ changes little; at $\lambda=0.3$, $J=1$ lies between $J=0$ and $J=2$. Where the selfgen excess is below 20\%, the running average of the logit range changes it by at most \runAvgMaxDiff\ points, consistent with \citet{skliar2025cache}, who find their running average comparable to a calibration-set estimate; at larger $\lambda$ the two differ more, far from the frontier. At 50\% residency every setting of \cp as published is dominated by a \psr configuration that is faster and closer at once. At 25\% residency, $J=2$ with $\lambda\ge0.5$ is faster than every \psr configuration we ran (\msJtwoMaxB\ ms at $\lambda=1$), but at a selfgen excess of \sgJtwoHalfB--\sgJtwoMaxB\%, far beyond the range in which any configuration kept its benchmark losses small; below that range every setting is dominated by \psr.

\textbf{What to prefetch.} Prefetching by the A priority of the next token (token prior and moving averages) gives exact routing little and hurts under \cp: at $\lambda=0.2$ and 50\% residency it raises latency from \msCPAnopfA\ to \msCPAthA\ ms, because the bias already keeps routing on resident experts and most prefetches are wasted. Cross-layer pre-gate candidates help because the router ranks them highly.

\textbf{Timing-dependent prefetch.} Throttling prefetches by transfer completion and letting routing see physical residency (\cref{sec:semantics}) gives about the same latency--perplexity trade-off as the logical variant (perplexity on the 12 timing windows). At 50\% residency, the timing-dependent configurations closest to the frontier reach \pplTdepA\ at \msTdepA\ ms and \pplTdepLoA\ at \msTdepLoA\ ms, while \psr reaches \pplPsrFifA\ at \msPsrFifA\ ms ($\lambda=0.15$, $n=1$) and \pplPsrNtwoA\ at \msPsrNtwoA\ ms ($\lambda=0.2$, $n=2$). Determinism therefore costs little or no speed.

\subsection{Speculative execution}
\label{sec:eval-spec}

Speculative execution runs the predicted experts whatever the cache holds, so its outputs, and therefore its quality, do not depend on residency. Its selfgen excess is \sgSpec\%, about the same as \psr at $\lambda=0.3$, $n=1$, 50\% residency (\sgFastA\%). It runs at \msSpecA\ ms (p99 \tailSpecA) at 50\% residency, \spdSpecA$\times$ slower than \psr (\msFastA\ ms, p99 \tailFastA), and at \msSpecB\ ms at 25\%, about as slow as the best exact-routing configuration (\msExactBestB\ ms).

On the benchmarks it loses to exact routing on MMLU ($-1.3$) and HellaSwag ($-0.9$), both significant after correction, and nominally on BBH ($-1.9$, $p=0.03$ before correction), with a mean of \meanExSpec\ $\pm$\meanExSpecci\ over the eight tasks (\cref{tab:bench}). In direct comparisons (\cref{tab:pairs}), \psr at \msFastA\ ms is not significantly worse on any task (mean \pairSpecA\ $\pm$\pairSpecAci), and the slower \psr point ($\lambda=0.2$, $n=1$, \msQualA\ ms) is nominally better on MMLU, HellaSwag and BBH, none of them significant after correction. At 25\% residency, \psr at \msFastB\ ms and speculative execution at \msSpecB\ ms are within noise of each other on every task (mean \pairSpecB\ $\pm$\pairSpecBci). Speculative execution therefore costs about as much quality as \psr while running about twice as slow at both residencies; \cref{sec:eval-why} explains why.

\subsection{A second model: Moonlight-16B-A3B}
\label{sec:moonlight}

\textbf{Model and port.} Moonlight-16B-A3B-Instruct~\citep{liu2025muon} uses the DeepSeek-V3 architecture~\citep{deepseekai2024v3}: 26 MoE layers with 64 routed experts each, top-6 routing by a sigmoid router with a selection bias, and two shared experts per layer. A routed expert is 17.3\,MB in bf16, 1.4 times an OLMoE expert, and the whole model takes 31\,GB, so on the 24\,GB RTX~4090 the routed experts must be offloaded. The engine keeps attention, the dense first layer, the shared experts and the routers on the GPU and the routed experts in pinned host memory; the harness streams them from host memory as well. The routing bonus of \cref{eq:cp} is added to the router's selection scores (sigmoid plus bias), and the selected experts are weighted with the unmodified scores. The A priority, its decays and the budget $n=1$ are those of OLMoE; the token prior, warm set and $\bar\Delta_l$ are estimated on held-out text as before. The one setting we chose on Moonlight is the pre-gate window. With OLMoE's 8 candidates, \psr ($\lambda=0.2$) makes \xferMoonPsrWide\ copies per token and saturates the link. Restricting insertion to the first 4 candidates reduces this to \xferMoonPsr\ copies at the cost of slightly more divergence, and we use this window below.

\textbf{Measurement.} Latency is measured as for OLMoE (\cref{sec:setup}) with the logical cache: means over \nRepMoon\ rounds in which the configurations of a cache size are interleaved. The 95\% intervals are within \moonCIpsrpct\% of the mean for \psr and up to \moonCIpct\% for exact routing with LRU, whose demand copies are the most exposed to host load. As the divergence diagnostic we use the KL divergence from exact routing on 132 windows of WikiText, C4 and self-generated text in the model's chat template. The benchmarks follow the OLMoE protocol, except that the harness decodes in batches of 8 on two GPUs (\cref{app:protocol}).

\begin{figure*}[t]
\centering
\includegraphics[width=\textwidth]{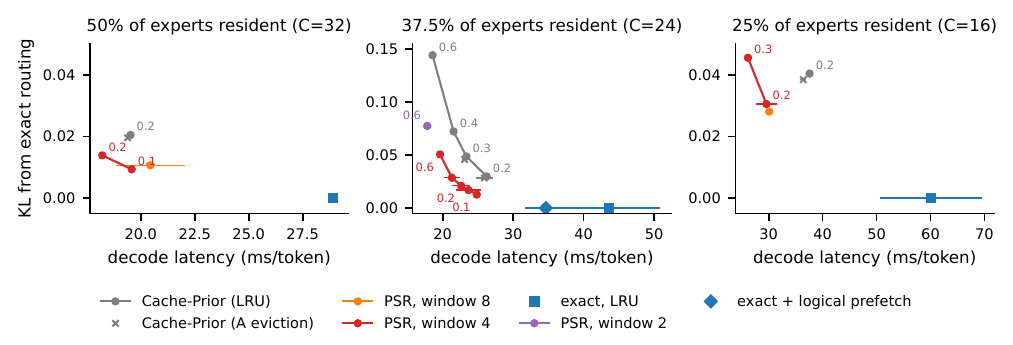}
\caption{Moonlight-16B-A3B: latency vs.\ KL divergence from exact routing at three cache sizes. \psr uses $n=1$ and a pre-gate window of 4 candidates (lines), 8 or 2 (single points); labels give $\lambda$. Exact routing with logical prefetch is \psr's prefetching without the bonus. Means over \nRepMoon\ interleaved rounds with 95\% intervals. Off exact routing, every point on the latency--KL Pareto front is a \psr configuration.}
\label{fig:moonlight}
\end{figure*}

\begin{table*}[t]
\caption{Moonlight-16B-A3B with $C=24$ of 64 experts resident (37.5\%). ms = mean engine latency per token over \nRepMoon\ rounds; copies = expert copies per token; KL from exact routing. Benchmark columns as in \cref{tab:bench}: the first row gives the absolute scores of exact routing, the other rows the paired difference to it (\textbf{bold}: significant after Holm's correction over the table's \nTestsMoon\ tests; $^\dagger$: $p<0.05$ before correction only); the last two rows compare \psr $\lambda=0.2$ directly with the two \cp configurations at about the same latency (iso-lat.) and the closest KL (sim.\ KL), with the mean $\pm$ 95\% interval. \psr uses the A priority, $n=1$ and a pre-gate window of 4.}
\label{tab:moonlight}
\centering
\small
\setlength{\tabcolsep}{2.6pt}
\begin{tabular}{@{}lrrrrrrrrrrrrr@{}}
\toprule
config & ms & copies & KL & MMLU & ARC-C & HellaS. & GSM8K & BBH & IFE-P & IFE-I & HumanE. & MBPP & mean \\
\midrule
exact, LRU & 43.6 & 36.7 & 0.000 & 67.9 & 57.5 & 62.2 & 74.5 & 66.6 & 44.2 & 56.0 & 58.5 & 57.8 & -- \\
exact + prefetch & 34.6 & 41.3 & 0.000 & \multicolumn{10}{c}{\emph{outputs identical to exact routing}} \\
\midrule
CP-A $\lambda$0.2 (A eviction) & 25.9 & 13.4 & 0.028 & \textbf{$-$0.6} & $-$0.7 & $-$0.3$^\dagger$ & +0.8 & $-$2.1$^\dagger$ & $-$1.7 & $-$1.8 & $-$4.3 & $-$2.2 & $-$1.37 \\
CP $\lambda$0.4 (LRU) & 21.5 & 8.0 & 0.072 & \textbf{$-$1.1} & $-$2.0$^\dagger$ & \textbf{$-$0.8} & $-$1.6 & \textbf{$-$3.7} & $-$5.5$^\dagger$ & \textbf{$-$5.9} & $-$1.2 & $-$0.2 & $-$2.03 \\
\midrule
PSR $\lambda$0.3 & 21.3 & 16.8 & 0.029 & $-$0.6$^\dagger$ & $-$1.7$^\dagger$ & $-$0.4$^\dagger$ & +0.6 & $-$1.7$^\dagger$ & $-$0.9 & $-$1.1 & $-$2.4 & $-$1.0 & $-$1.03 \\
PSR $\lambda$0.2 & 22.6 & 18.5 & 0.021 & $-$0.3 & $-$0.9 & $-$0.3$^\dagger$ & +1.8 & $-$1.5$^\dagger$ & $-$0.9 & $-$0.6 & $-$1.2 & $-$0.6 & $-$0.48 \\
\midrule
vs.\ CP $\lambda$0.4 (iso-lat.) & \multicolumn{3}{c}{22.6 vs.\ 21.5 ms} & \textbf{+0.9} & +1.2 & \textbf{+0.6} & +3.4$^\dagger$ & +2.1$^\dagger$ & +4.6$^\dagger$ & \textbf{+5.3} & 0.0 & $-$0.4 & +1.6$\pm$0.9 \\
vs.\ CP-A $\lambda$0.2 (sim.\ KL) & \multicolumn{3}{c}{22.6 vs.\ 25.9 ms} & +0.4$^\dagger$ & $-$0.2 & 0.0 & +1.0 & +0.6 & +0.7 & +1.2 & +3.0 & +1.6 & +0.9$\pm$0.8 \\
\bottomrule
\end{tabular}

\end{table*}

\textbf{The ordering of \cref{sec:eval-latency,sec:eval-quality} holds.} \Cref{fig:moonlight} shows the latency--KL trade-off at 50\%, 37.5\% and 25\% residency. At every cache size, every configuration on the Pareto front other than exact routing is a \psr configuration; no \cp configuration, with LRU or with the A priority, is on it. This includes the high-$\lambda$ end, where \cp nearly freezes the cache: at 37.5\% residency \psr with $\lambda=0.6$ and a window of 2 is faster than \cp with $\lambda=0.6$ (\msMoonPsrSix\ against \msMoonCPsix\ ms) at about half its KL (\klMoonPsrSix\ against \klMoonCPsix). At 50\% residency \psr ($\lambda=0.2$) runs at \msMoonPsrHalf\ ms with KL \klMoonPsrHalf, against \msMoonCPAHalf\ ms and \klMoonCPAHalf\ for \cp with the A priority and \msMoonExactHalf\ ms for exact routing with LRU. At 25\% residency the same comparison gives \msMoonPsrQuarter\ ms (KL \klMoonPsrQuarter), \msMoonCPAQuarter\ ms (\klMoonCPAQuarter) and \msMoonExactQuarter\ ms.

We ran the benchmarks at 37.5\% residency (\cref{tab:moonlight}). \psr with $\lambda=0.2$ decodes at \msMoonPsr\ ms/token (p99 \tailMoonPsr\ ms), \spdMoonExact$\times$ faster than exact routing with LRU (\msMoonExact\ ms, p99 \tailMoonExact) and faster than exact routing with logical prefetch (\msMoonPrefetch\ ms). \psr at 37.5\% residency is also faster than exact routing at 50\% (\msMoonExactHalf\ ms), with a quarter less expert memory.
\begin{itemize}
  \item \textbf{Same latency.} \cp needs $\lambda=0.4$ to reach \psr's latency (\msMoonCPfour\ ms, p99 \tailMoonCPfour, against \msMoonPsr\ ms), and its KL is then \klMoonCPfour\ against \psr's \klMoonPsr. On the benchmarks, \psr is ahead by \pairMoonIsoLat\ $\pm$\pairMoonIsoLatci\ points on average. After Holm's correction over the \nTestsMoon\ tests of \cref{tab:moonlight}, its gains on MMLU, HellaSwag and IFEval at instruction level ($+5.3$) are significant; those on GSM8K ($+3.4$), BBH ($+2.1$) and IFEval at prompt level ($+4.6$) are nominal ($p<0.05$ before correction); the code tasks tie.
  \item \textbf{Similar KL.} \cp with the A priority and $\lambda=0.2$ has a KL of \klMoonCPA, close to \psr's, and runs at \msMoonCPA\ ms. \psr has \redMoonCPA\% lower latency, and its scores are lower only on ARC-C ($-0.2$); no task differs after correction, and the mean is \pairMoonIsoQual\ $\pm$\pairMoonIsoQualci.
  \item \textbf{Against exact routing}, \psr $\lambda=0.2$ loses \meanExMoonPsr\ $\pm$\meanExMoonPsrci\ points on average, with no significant loss after correction (nominal on HellaSwag, $-0.3$, and BBH, $-1.5$). Raising $\lambda$ to 0.3 saves \redMoonHi\% of latency and doubles the mean loss (\meanExMoonPsrHi).
\end{itemize}
Across the four configurations, the ranking by KL (\klMoonPsr\ $<$ \klMoonCPA\ $\approx$ \klMoonPsrHi\ $<$ \klMoonCPfour) matches the ranking by mean benchmark loss (\meanExMoonPsr; \meanExMoonCPA\ and \meanExMoonPsrHi; \meanExMoonCPfour). \cp at $\lambda=0.4$ loses most on instruction following and long chain-of-thought (IFEval $-5.5$, BBH $-3.7$), not on code.

\subsection{A third model: Qwen3.6-35B-A3B}
\label{sec:qwen}

\textbf{Model and port.} Qwen3.6-35B-A3B~\citep{qwen36_35b_a3b} has 40 layers, 30 with Gated DeltaNet~\citep{yang2025gated} and 10 with gated attention as token mixer, each followed by an MoE block with 256 routed experts, top-8 softmax routing and one gated shared expert. We use the official FP8 checkpoint (block-wise e4m3 weights) as the reference model, so exact routing is that checkpoint's own. A routed expert takes 3.2\,MB, a quarter of an OLMoE expert, and one copy about 0.14\,ms. The routed experts take 33\,GB, so their offloading is real. The engine keeps the structure of \cref{sec:impl}, two CUDA graphs per layer with the routing policy on the host between them, and runs the FP8 layers on vLLM's kernels. The harness runs the model on two GPUs. \psr's settings are those of Moonlight: A priority, $n=1$ and a pre-gate window of 4 candidates (we also time a window of 2). The token prior, warm set and $\bar\Delta_l$ are estimated on held-out text as before. We fixed the two benchmark configurations before any benchmark result: \cp at its operating point $\lambda=0.2$ and \psr with $\lambda=0.2$, both at 25\% residency ($C=64$).

\textbf{Engine and harness agree in perplexity, not in expert sets.} The check of \cref{sec:impl}, agreement of selected expert sets per token and layer, is not informative here. The router's logit margin between its 8th and 9th expert is below 0.01 on 18\% of token-layers in bf16, and the FP8 quantization of activations turns one-ulp differences into different selections. Under exact routing, transformers' own incremental decoding and full forward pass already select different sets on 31\% of token-layers. We therefore compare perplexity on 17 windows: under exact routing the engine is within \valQwPpl\% of the harness, and \psr's change in negative log-likelihood from exact routing (\valQwEff\ nats per token in the harness at 25\% residency) agrees between the two to \valQwDD\ $\pm$\valQwDDci.

\textbf{Measurement.} Latency is measured as for Moonlight (\nRepQw\ interleaved rounds per cache size, the 12 teacher-forced windows of \cref{sec:setup}). The KL diagnostic uses every second one of Moonlight's 132 windows, re-tokenized. Host-side selection matters on this model: our first timing ran the \cp selection in numpy (116\,\textmu s per layer, 4.6\,ms per token over 40 layers), which made every cache-aware configuration slower than exact routing at 37.5\% and 50\% residency. All numbers below use a compiled selection (14\,\textmu s per layer) that selects the same experts. We ran a reduced benchmark suite (MMLU, ARC-C, HellaSwag, GSM8K, IFEval, HumanEval) and changed the GSM8K protocol (\cref{app:protocol}).

\begin{figure*}[t]
\centering
\includegraphics[width=\textwidth]{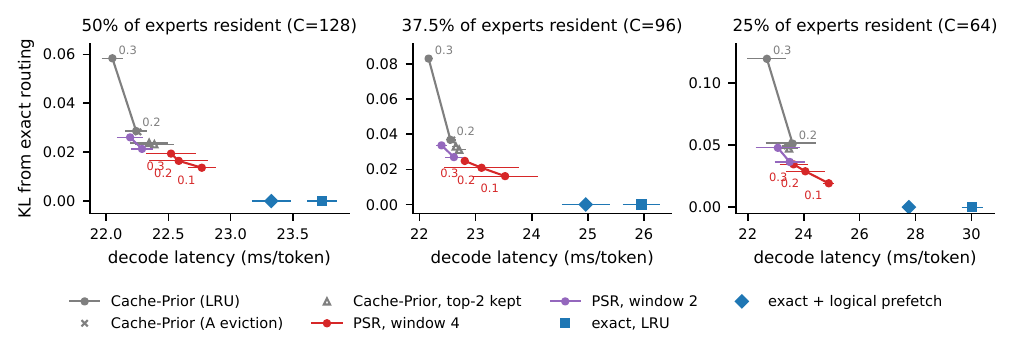}
\caption{Qwen3.6-35B-A3B (FP8): latency vs.\ KL divergence from exact routing at three cache sizes of 256 experts per layer. \psr uses $n=1$ and a pre-gate window of 4 or 2 candidates; labels give $\lambda$. Means over \nRepQw\ interleaved rounds with 95\% intervals. Every cache-aware configuration runs within \msQwSpread\,ms of the fastest; at the fast end of the Pareto front, it includes \cp configurations.}
\label{fig:qwen}
\end{figure*}

\begin{table*}[t]
\caption{Qwen3.6-35B-A3B (FP8) with $C=64$ of 256 experts resident (25\%). ms = mean engine latency per token over \nRepQw\ rounds; copies = expert copies per token; KL from exact routing. Benchmark columns as in \cref{tab:moonlight}, on a reduced suite of six tasks (\cref{app:protocol}): the first row gives the absolute scores of exact routing, the other rows the paired difference to it (\textbf{bold}: significant after Holm's correction over the table's \nTestsQwen\ tests; $^\dagger$: $p<0.05$ before correction only); mean over the six tasks with IFE-P. The last row compares \psr directly with \cp, with the mean $\pm$ 95\% interval. Both configurations were fixed before any benchmark was run.}
\label{tab:qwen}
\centering
\small
\setlength{\tabcolsep}{3pt}
\begin{tabular}{@{}lrrrrrrrrrrr@{}}
\toprule
config & ms & copies & KL & MMLU & ARC-C & HellaS. & GSM8K & IFE-P & IFE-I & HumanE. & mean \\
\midrule
exact, LRU & 30.0 & 53.5 & 0.000 & 83.6 & 54.6 & 63.4 & 95.5 & 83.5 & 88.7 & 58.5 & -- \\
exact + prefetch & 27.8 & 56.2 & 0.000 & \multicolumn{8}{c}{\emph{outputs identical to exact routing}} \\
\midrule
CP $\lambda$0.2 (LRU) & 23.6 & 8.3 & 0.051 & \textbf{$-$1.3} & $-$1.5 & \textbf{$-$0.8} & $-$0.3 & $-$0.4 & +0.1 & +3.0 & $-$0.21 \\
PSR $\lambda$0.2 & 24.1 & 16.5 & 0.029 & $-$0.4$^\dagger$ & $-$0.4 & $-$0.3 & $-$0.7 & $-$2.6 & $-$1.6 & +0.6 & $-$0.63 \\
\midrule
PSR vs.\ CP & \multicolumn{3}{c}{24.1 vs.\ 23.6 ms} & \textbf{+0.8} & +1.1 & \textbf{+0.6} & $-$0.4 & $-$2.2 & $-$1.7 & $-$2.4 & $-$0.4$\pm$1.0 \\
\bottomrule
\end{tabular}

\end{table*}

\textbf{Latency is bounded by computation.} \Cref{fig:qwen} shows the latency--KL trade-off at 50\%, 37.5\% and 25\% residency. The fastest configuration runs at \msQwFloor\ ms/token with \xferQwFloor\ copies per token, and every cache-aware configuration lies within \msQwSpread\,ms of it: \msQwRngHalf, \msQwRngMid\ and \msQwRngQuarter\ ms at the three cache sizes. Exact routing with LRU takes \msQwExactHalf, \msQwExactMid\ and \msQwExact\ ms, so \psr ($\lambda=0.2$) is only \spdQwExactHalf, \spdQwExactMid\ and \spdQwExactQuarter$\times$ faster. Copies are cheap relative to computation: exact routing at 25\% residency makes \xferQwExact\ copies per token, about 7.5\,ms of transfer if none overlapped, and once the routing bonus removes most of them, little is left for prefetching to hide. At 25\% residency \psr and \cp then run at the same speed within the intervals (\msQwPsr\ $\pm$\ciQwPsr\ and \msQwCP\ $\pm$\ciQwCP\ ms), and \psr's KL is \klQwRatio\ times \cp's (\klQwPsr\ vs.\ \klQwCP). Unlike on Moonlight, \cp configurations are on the Pareto front, at its fast end. \cp with $\lambda=0.3$ is the fastest point at every cache size, at more than twice the KL of the \psr points next to it (at 25\%: \msQwCPthree\ ms, KL \klQwCPthree). \cp with the router's top-2 kept is on the front at 25\% residency by a margin far inside the intervals: \psr with a window of 2 runs at the same latency (\msQwPsrTwo\ vs.\ \msQwCPJtwo\ ms) with lower KL (\klQwPsrTwo\ vs.\ \klQwCPJtwo). In greedy generation on 32 GSM8K and 32 HumanEval chat prompts (\cref{tab:qwensd}), \cp is \fgQwSlowRng\% faster than \psr at 25\% residency (\fgQwCPGsm\ vs.\ \fgQwPsrGsm\ ms on GSM8K, \fgQwCPHe\ vs.\ \fgQwPsrHe\ on HumanEval, in each of three rounds). As on OLMoE at 50\% residency (\cref{sec:engine-quality}), \cp needs few copies on the model's own outputs (\fgQwCPcpGsm\ and \fgQwCPcpHe\ per token), and \psr's prefetches (\fgQwPsrcpGsm\ and \fgQwPsrcpHe) cost more than they hide.

\textbf{\psr's lower KL shows on the likelihood tasks} (\cref{tab:qwen}). Against exact routing, \cp loses significantly after correction on MMLU (\dQwCPMmlu) and HellaSwag (\dQwCPHella). \psr has no significant loss; on MMLU it is nominal (\dQwPsrMmlu). In the direct comparison \psr is ahead on MMLU (\dQwPairMmlu) and HellaSwag (\dQwPairHella), both significant after correction. On the generative tasks no difference is significant even before correction ($p\ge\pQwPairGenMin$), and the point estimates favour \cp (IFEval \dQwPairIfe, HumanEval \dQwPairHe). Averaged over the six tasks the two configurations do not differ (\pairQw\ $\pm$\pairQwci). The likelihood tasks are scored on 10--14 thousand questions and resolve differences well below a point; the generative ones resolve differences of 1--5 points. At that resolution the generative tasks do not register the routing changes of either policy, although \cp replaces more of the router's choices (\dispQwCP\ of 320 router slots per token across the benchmarks, against \dispQwPsr\ for \psr).

On this model, then, \psr keeps its advantage in divergence and in the precise benchmarks over \cp at equal teacher-forced latency, but not its speed: decoding is bounded by computation, and in free generation \cp is slightly faster. The mechanism of \cref{sec:eval-why}, hiding copies behind computation, pays only where copies are a large share of the time per token.

\subsection{Does training the gate help?}
\label{sec:training}

Before settling on training-free routing we trained the gate, which is a natural way to reduce the deviation that cache-aware routing causes. We trained a gate that keeps a persistent active set and swaps an expert in only when its logit exceeds the weakest member's by a margin; with a uniform margin for resident experts this rule is \cref{eq:cp} on a model-defined cache, which we verified numerically. Training distils the frozen model (KL on the next-token distribution) with a smooth penalty on host loads, learns a per-token margin from the gate's state, and adapts the router and LoRA adapters~\citep{hu2022lora} (rank 16 on attention, 8 per expert) on 4.5M tokens.

Training lowered perplexity substantially at the same number of host loads (selfgen excess about 17--30\% lower than \cp), but not benchmark quality. At about ten host loads per token and 50\% residency, the trained gate and \cp are within noise of each other on BBH, IFEval and the code tasks (e.g.\ BBH $+0.2$, HumanEval $-4.3$ with $p=0.09$). The trained gate is nominally better on GSM8K ($+2.9$, $p=0.02$), but the gain comes from the LoRA adapters: with only the router trained it disappears ($-0.3$). Both trained variants are nominally worse than \cp on MMLU ($-0.5$ and $-0.6$, $p=0.05$ and $0.02$), and the router-only variant is better on HellaSwag ($+0.8$, $p<0.001$), the only one of these differences that is significant after correction over the nine tasks of each comparison. In short, the extra deviation that training removes is not deviation the benchmarks detect.

We repeated the experiment with \psr itself, on OLMoE at 37.5\% residency ($C=24$, $\lambda=0.3$, $n=1$). The router, attention LoRA (rank 16) and expert LoRA (rank 8) are distilled to the exact model through \psr's selection on 3.9M tokens. At matched copies this lowers the selfgen excess by about 20\%, but on the benchmarks it is a trade, not a gain. Against the untrained rule at the same $\lambda$ and copies, BBH improves by \trBBH\ ($p=\trBBHp$), which offsets the untrained rule's BBH loss to exact routing (\trBBHexUn, $p=\trBBHexUnp$), while HellaSwag drops by \trHella\ ($p<0.001$). Only the HellaSwag drop is significant after correction over the nine tasks of the comparison; both BBH differences are nominal. The eight-task mean difference is \trMean\ $\pm$\trMeanci. A learned, bounded per-token bonus was the only training variant that improved the selfgen frontier further (about 8\% at matched copies). It is likewise within noise of the untrained rule at a lower $\lambda$ with about the same copies (\trBonusMean\ $\pm$\trBonusMeanci). With this data and objective (KL to the exact model), training improves the diagnostic but not the benchmarks; other data or rewards might change this. Methods that fine-tune for expert reuse report gains~\citep{raje2026melinoe,zhu2026remoe,wu2026cacheaware}; we evaluate against a training-free cache-aware baseline at matched loads and with generative benchmarks, which may explain the difference.

\begin{table*}[t]
\caption{Speculative decoding (SD) on OLMoE with $P=24$ experts per layer (37.5\%), RTX~4090, batch 1. pass@1 and strict: full benchmarks (164 and 1319 problems) in the harness. $\tau$ = committed tokens per verification, copies = expert copies per generated token, ms = decode latency per generated token: all three from the engines on the timed subset (the first 80 HumanEval and 48 GSM8K problems). The copy ratios quoted in \cref{sec:sd} use the copies of the full benchmark runs instead, which have longer accepted runs on HumanEval. AR-PSR = \psr with $C=24$ and $n=1$. SD rows use $k=16$ drafts from OLMo-1B unless noted. Position weights are AcceptMoE's; confidence + top-1 is our variant (\cref{sec:sd}). ``Copies skipped'' runs the same routing without any PCIe copy and gives the compute-only time; it and $k=8$ were only timed. --: not measured.}
\label{tab:sd}
\centering
\small
\begin{tabular}{@{}lrrrrrrrr@{}}
\toprule
 & \multicolumn{4}{c}{HumanEval} & \multicolumn{4}{c}{GSM8K} \\
\cmidrule(lr){2-5}\cmidrule(lr){6-9}
config & pass@1 & $\tau$ & copies & ms & strict & $\tau$ & copies & ms \\
\midrule
AR, all experts resident (exact routing) & 32.9 & -- & 0.0 & 6.1 & 33.2 & -- & 0.0 & 6.5 \\
\midrule
AR-PSR $\lambda$0.25 & 30.5 & -- & 10.0 & 9.2 & 32.8 & -- & 15.2 & 11.3 \\
AR-PSR $\lambda$0.3 & 34.8 & -- & 9.8 & 9.1 & 33.4 & -- & 14.1 & 10.5 \\
\midrule
SD, exact verification & 31.1 & 6.5 & 29.3 & 27.3 & 33.1 & 4.9 & 59.7 & 53.2 \\
SD, block selection, position weights & 35.4 & 6.7 & 6.2 & 13.6 & 34.3 & & & -- \\
SD, block selection, confidence + top-1 & 29.9 & 6.4 & 6.4 & 14.3 & 33.5 & 4.5 & 16.4 & 26.3 \\
\quad same, copies skipped & -- & 6.4 & -- & 10.7 & -- & 4.5 & -- & 16.4 \\
\quad same, $k=8$ & -- & 5.0 & 9.3 & 13.1 & -- & 4.2 & 21.4 & 22.5 \\
\bottomrule
\end{tabular}

\end{table*}

\section{Speculative Decoding with Block-Level Expert Selection}
\label{sec:sd}

Speculative decoding~\citep{leviathan2023fast,chen2023accelerating} lets a small drafter propose $k$ tokens that the target model verifies in one forward pass of $k+1$ positions. Under offloading, verification is expensive, because each layer needs the union of the experts of all positions: about 40 of OLMoE's 64 experts per layer at $k=16$. With exact routing, speculative decoding therefore copies more expert bytes per committed token than autoregressive (AR) decoding. AcceptMoE~\citep{liang2026acceptmoe} restricts verification to one eligible expert set per block and layer. The set is chosen from the router's demand at all positions, weighted by how often a draft at that position is committed, and pruned toward resident experts; every position then routes within the set. We ask whether such block-level selection beats \psr at batch 1 once the drafter and the verification are paid for.

\textbf{Setup.} The target is OLMoE with a pool of $P=24$ experts per layer (37.5\%) and LRU replacement; the drafter is OLMo-1B-0724~\citep{groeneveld2024olmo}, which shares OLMoE's tokenizer, with greedy decoding. Besides AcceptMoE's position weights we test two changes. First, we weight each position's demand by the drafter's cumulative confidence in the draft up to that position, $\alpha_d=(\prod_{i\le d}c_i)^{1/2}$, where $c_i$ is the drafter's probability of its own token. Rejected drafts are wrong tokens, and their router demand pulls experts into the set that no committed token uses. Second, each position whose weight $\alpha_d$ is at least 0.3, i.e.\ each draft likely to be committed, keeps its own top-1 expert, even if it lies outside the set. Without this rescue, restricting positions to the set displaces whole top-1 experts, and the KL from exact routing is about twice that of \psr at the same displaced router mass. The speculative decoding engine replays CUDA graphs for the drafter and, per layer, for the $(k+1)$-position verification. Like the AR engine, it uses the FlashAttention-2 kernel of vLLM. Expert selection runs on the host, and the copies of missing experts are on a copy stream that the layer waits on. Scores come from the full benchmarks in the HuggingFace harness. Latency is measured with the engines on the first 80 HumanEval and 48 GSM8K problems, decoding only, against \psr ($C=24$, $n=1$) on the same prompts. On the timed HumanEval problems the engines reproduce the benchmark scores of the harness (10 problems solved only by the engines, 12 only by the harness). Against a full forward pass, their next-token distributions deviate as much as the harness's own incremental decoding does (KL about 0.002, top-1 agreement 98.8\%).

\begin{table*}[t]
\caption{Qwen3.6-35B-A3B (FP8): speculative decoding (SD) with the DFlash block drafter against AR decoding, batch 1, on the first 32 GSM8K and 32 HumanEval problems as single chat prompts (thinking disabled, greedy, at most 512 new tokens); means over 3 rounds. AR rows at $C=64$ (25\%), $C=128$ (50\%, the expert slots of SD) and $C=\CQwMatch$ (the largest AR capacity within SD's peak GPU memory, drafter included); rows at $C\ge128$ and the compute-only AR row (no copies, forced along exact routing's continuation) come from three further rounds. SD rows: a pool of $P=64$ experts per layer between blocks plus 64 slots for a block's union, $k=15$. ms = decode latency per generated token; copies = expert copies per generated token; $\tau$ = committed tokens per verification; strict and pass@1 = scores of these generations (identical in every round). ``Copies skipped'' runs the same selection without any PCIe copy and gives the compute-only time; its outputs are not meaningful.}
\label{tab:qwensd}
\centering
\small
\begin{tabular}{@{}lrrrrrrrr@{}}
\toprule
 & \multicolumn{4}{c}{GSM8K (32 prompts)} & \multicolumn{4}{c}{HumanEval (32 prompts)} \\
\cmidrule(lr){2-5}\cmidrule(l){6-9}
config & ms & copies & $\tau$ & strict & ms & copies & $\tau$ & pass@1 \\
\midrule
AR $C{=}64$, exact routing, LRU & 36.3 & 85.0 & -- & 78.1 & 36.0 & 82.6 & -- & 100.0 \\
AR $C{=}64$, CP $\lambda$0.2 & 24.8 & 9.2 & -- & 75.0 & 24.3 & 6.4 & -- & 100.0 \\
AR $C{=}64$, PSR $\lambda$0.2 & 25.8 & 24.7 & -- & 81.2 & 25.7 & 23.9 & -- & 100.0 \\
\midrule
AR $C{=}128$, exact routing, LRU & 27.9 & 28.7 & -- & 78.1 & 28.0 & 30.0 & -- & 100.0 \\
AR $C{=}128$, CP $\lambda$0.2 & 24.4 & 2.5 & -- & 84.4 & 24.2 & 1.8 & -- & 100.0 \\
AR $C{=}128$, CP $\lambda$0.3 & 24.0 & 1.2 & -- & 59.4 & 24.0 & 0.8 & -- & 100.0 \\
AR $C{=}128$, PSR $\lambda$0.2 & 25.0 & 10.1 & -- & 84.4 & 25.0 & 11.0 & -- & 100.0 \\
AR $C{=}148$, CP $\lambda$0.2 & 24.2 & 1.8 & -- & 75.0 & 24.1 & 1.3 & -- & 100.0 \\
AR $C{=}148$, PSR $\lambda$0.2 & 24.8 & 7.9 & -- & 81.2 & 24.8 & 9.0 & -- & 100.0 \\
AR, copies skipped & 22.9 & 0 & -- & -- & 23.0 & 0 & -- & -- \\
\midrule
SD, exact verification & 23.0 & 105.2 & 7.8 & 78.1 & 15.8 & 76.0 & 12.8 & 100.0 \\
SD, block selection, confidence + top-1 & 11.2 & 27.1 & 7.7 & 75.0 & 8.4 & 26.6 & 13.0 & 100.0 \\
\quad same, copies skipped & 7.0 & 0 & -- & -- & 5.1 & 0 & -- & -- \\
\bottomrule
\end{tabular}

\end{table*}

\textbf{Bytes.} \Cref{tab:sd} shows that block-level selection cuts the copies of exact speculative decoding by a factor of 3.6--5. Compared with \psr, its value grows with the acceptance length $\tau$. On the full HumanEval benchmark (about 7 committed tokens per verification), our variant needs \sdBytesConfHE$\times$ the copies per token of \psr $\lambda=0.25$ and AcceptMoE's weighting \sdBytesPosHE$\times$. On GSM8K with 5-shot prompts (about 5 tokens per verification), the ratios are \sdBytesConfGSM$\times$ and \sdBytesPosGSM$\times$. No score difference is resolvable. In paired tests among the seven benchmarked configurations (those of \cref{tab:sd} and confidence weighting with a top-1 rescue at every position), 2 of 21 HumanEval pairs reach $p<0.05$, about what chance gives, and no GSM8K pair does ($p\ge0.24$).

\textbf{Time.} Speculative decoding is nonetheless slower than \psr. Its fastest timed configuration, our variant with $k=8$ (timed only, not benchmarked), takes \sdRatioHE$\times$ (HumanEval) and \sdRatioGSM$\times$ (GSM8K) the time per token of \psr. Even without any copies it would take \sdFreeHE$\times$ and \sdFreeGSM$\times$ at $k=16$. The time per verification with $k=16$ decomposes as follows.
\begin{itemize}
  \item \textbf{Drafting dominates.} One OLMo-1B step takes \sdDraftStep\ ms, more than half of a full target step (\msArFloorHE--\msArFloorGSM\ ms with all experts resident): OLMoE activates 1.3B parameters per token and OLMo-1B has 1.2B. Drafting 16 tokens therefore takes \sdDraftHE--\sdDraftGSM\ ms, or \sdDraftTokHE\ and \sdDraftTokGSM\ ms per committed token, about as much as a whole \psr token (\msArPsrHE\ and \msArPsrGSM\ ms). A smaller $k$ halves the drafting per verification, but $\tau$ falls and the copies per token rise.
  \item \textbf{Verification compute is cheap.} With copies skipped, verifying 17 positions takes \sdVerCompHE--\sdVerCompGSM\ ms, under two target steps, because a batch-1 forward pass is bound by memory bandwidth.
  \item \textbf{Verification copies are exposed.} Each copy adds \sdCopyMsHE--\sdCopyMsGSM\ ms, about its full transfer time: the engine chooses a layer's eligible set from that layer's router outputs, known only after its attention, and does not prefetch. \psr's cross-layer prefetch brings its effective cost per copy down to \arCopyEffHE--\arCopyEffGSM\ ms. Verification alone takes \sdVerRatioHE$\times$ (HumanEval) and \sdVerRatioGSM$\times$ (GSM8K) of \psr's time per token, close to the byte ratios.
\end{itemize}
Holding $\tau$ and the copies fixed, speculative decoding would match \psr on HumanEval if drafting 16 tokens cost at most \sdBEHE\ ms, \sdBEstepsHE\ target steps. On GSM8K, verification alone is slower than \psr, so even free drafting does not suffice. If verification copies were hidden as well as \psr hides its own, the break-even drafting time would be \sdBEhidHE\ ms (\sdBEhidStepsHE\ steps) on HumanEval and \sdBEhidGSM\ ms (\sdBEhidStepsGSM\ steps) on GSM8K. Drafters that produce a block in one forward pass~\citep{chen2026dflash} or reuse the target's features~\citep{li2024eagle} may fall in this range. None exists for OLMoE, and its target-to-drafter cost ratio is unusually small; we test a block drafter on a larger model below.

\textbf{A block drafter on Qwen3.6.} DFlash~\citep{chen2026dflash} provides a drafter for Qwen3.6-35B-A3B (\cref{sec:qwen}): a 6-layer block-diffusion model that conditions on hidden states of eight target layers and drafts $k=15$ tokens in one forward pass. We ported the speculative decoding engine to Qwen with the same block selection (confidence weights, top-1 rescue) and timed it against AR decoding with the same 32 GSM8K and 32 HumanEval chat prompts and decoding protocol (\cref{tab:qwensd}). SD holds $P=64$ experts per layer between blocks plus 64 slots for a block's union, the expert slots of AR at $C=128$; with the drafter it peaks at \memQwSd\ GB of GPU memory, against \memQwArHalf\ GB for AR at $C=128$ and \memQwArMatch\ GB at $C=\CQwMatch$, the largest AR capacity within SD's peak. One drafting pass takes about \sdQwDraftBlkGsm\ ms, under half a target step (OLMo-1B needed 16 sequential steps of more than half a target step each); per committed token, drafting costs \sdQwDraftGsm\ and \sdQwDraftHe\ ms. With \sdQwTauGsm\ and \sdQwTauHe\ committed tokens per verification, block selection decodes at \sdQwConfGsm\ and \sdQwConfHe\ ms per token on GSM8K and HumanEval: \spdQwSdExactRng$\times$ faster than AR with exact routing at the same expert slots and \spdQwSdCPRng$\times$ faster than the fastest AR configuration (\cp $\lambda=0.3$, $C=128$; 95\% intervals over prompts \spdQwSdBootGsmci\ and \spdQwSdBootHeci, faster on every prompt). AR decoding is bounded by computation (\cref{sec:qwen}): its cache-aware configurations take \arQwHalfRng\ ms at $C=128$ and \arQwMatchRng\ ms at $C=\CQwMatch$, and \arQwFloorGsm\ and \arQwFloorHe\ ms with every copy skipped (forced along exact routing's continuation). No residency can make AR decoding in our engine faster than this on these prompts; block selection is \spdQwSdFloorRng$\times$ faster. Faster kernels would lower this bound, but would also speed up verification. Exact verification, whose routing is exact up to numerics, runs at \sdQwExactGsm\ and \sdQwExactHe\ ms; the fastest AR configuration takes \spdQwSdExactVerGsm\ and \spdQwSdExactVerHe\ times as long. Copies remain exposed as on OLMoE: without them, block selection would take \sdQwFreeGsm\ and \sdQwFreeHe\ ms (with copies, re-timed in the rounds of the $C\ge128$ rows: \sdQwConfBGsm\ and \sdQwConfBHe\ ms). Of the 32 GSM8K problems, block selection solves \scQwSdConfGsm\%, AR with exact routing \scQwArExactGsm\% and the fastest AR configuration \scQwArFastGsm\%; every configuration solves all 32 HumanEval problems. Such samples detect only large losses, and we did not run the benchmarks with block selection on Qwen. With a drafter this cheap, the break-even of the OLMoE analysis is met: speculative decoding with block-level selection is the fastest configuration we measured on Qwen, and exact verification alone is at least as fast as cache-aware AR routing.

\textbf{The routed model now includes the drafter.} Block-level selection makes the experts executed for a committed token depend on the other drafts in its block. In a replay of 3346 recorded verification blocks with AcceptMoE's selection, 36\% of committed token-layers execute a different expert set when the rejected positions carry the drafter's wrong tokens than when they carry the true continuation. The output of such a system therefore depends on the drafter and on $k$, and differs from that of AR decoding with the same cache-aware rule. It is still deterministic: the pool is a function of the token history and the drafts, so a harness can reproduce it, which is the property \cref{sec:semantics} requires. A per-position logical cache with rollback on rejection would keep speculative decoding's output identical to \psr's. In a CPU replay with a synthetic drafter, however, it needed 1.4--4$\times$ the copies of block-level selection at the same displaced router mass, and at $k=16$ more copies per token than \psr itself.

\section{Related Work}
\label{sec:related}

\textbf{Offloading, caching and prefetching with exact routing.} \citet{eliseev2023fast} combine an LRU expert cache with speculative loading for Mixtral. MoE-Infinity~\citep{xue2024moeinfinity} uses activation-aware caching on personal machines. ProMoE~\citep{song2024promoe} learns to predict upcoming experts for proactive caching, and Pre-gated MoE~\citep{hwang2024pregated} fine-tunes the model so that each layer's experts are selected one layer early, which changes the model but keeps it deterministic; ExpertFlow~\citep{shen2025expertflow} adapts the prediction horizon to bandwidth and schedules tokens whose experts are resident first, without changing which experts run. HybriMoE~\citep{zhong2025hybrimoe} schedules experts across CPU and GPU, prefetches by estimated impact, and replaces cached experts by a moving average of their router scores, including high-scoring experts that were not selected, a replacement rule close to our A priority. Further work refines prediction and replacement~\citep{wang2026seqmoe,kim2026flashmoe,yang2026moecore}, adapts execution to bandwidth~\citep{yang2026freetoken}, or computes missing experts on the CPU instead of copying them~\citep{kamahori2024fiddler,chen2025ktransformers}. MoE-Lightning~\citep{cao2025moelightning} pipelines CPU, GPU and I/O for high-throughput batched inference under memory limits, a different regime from our batch-1 latency. HOBBIT~\citep{tang2024hobbit} loads low-precision copies of less important missed experts. These systems are complementary to \psr; our exact-routing baselines include the cross-layer prediction and score-based replacement they use.

\textbf{Changing which experts run.} \Cref{tab:mechanisms} compares the closest methods along three design decisions: how future experts are predicted, whether the prediction changes the executed set, and which cache state routing reads. \citet{skliar2025cache} introduced the routing bonus we build on (\cp), for flash-to-DRAM offloading on mobile devices, with an LRU cache and no prefetching; MoE-ERAS~\citep{bambhaniya2024eras} also prefers resident experts. SpecMD~\citep{hoang2026specmd} combines this bonus with cross-layer prefetching, eviction policies and miss handling in a common framework, evaluated with emulated bandwidth and quantized experts; the bonus applies to the GPU-resident set, so routing depends on when prefetches land. SMoE~\citep{zhu2025smoe} substitutes low-score activated experts with resident experts of similar score, prefetches high-score experts one layer ahead, and falls back to the CPU; MoE-CORE~\citep{yang2026moecore} bounds cross-layer prefetching and optionally substitutes low-score misses. In these systems, too, the cache that routing or substitution reads depends on transfer timing and (for SMoE) batch composition; this is our reading, the authors do not discuss it. Speculating Experts~\citep{madan2026speculating} executes the predicted experts; on some models this costs substantial accuracy, which a trained estimator and a hybrid strategy largely recover. CommitMoE~\citep{li2026commitmoe} likewise commits execution to the experts predicted by a trained per-layer predictor, never loads the router's choice on a misprediction, and optionally redistributes the router's weights from missed to correctly predicted experts. Committing removes all corrective copies and keeps outputs independent of timing, but every misprediction changes the computation; our speculative-execution baseline uses the next layer's router as predictor and the true router weights, and is neither method in full. AdapMoE~\citep{zhong2024adapmoe} reduces the number of experts for insensitive tokens and combines this with prefetching and caching, but its gating does not depend on the cache. Expert skipping~\citep{lu2024notall} drops low-weight experts. \psr differs in choosing explicitly, under a budget and on a logical cache, which predicted experts become eligible for the bonus; the router still makes the final choice among resident and non-resident experts.

\begin{table}[t]
\caption{How methods that change the executed experts use predictions and cache state. Prediction: of layer $l$'s experts before layer $l$ runs. Executed: what decides the experts that run. Routing reads: the cache state that selection depends on (\emph{physical} = filled by asynchronous copies, hence timing-dependent; \emph{logical} = \psr's cache with at most $n$ insertions per layer and token). subst.\ = substitution of missed experts.}
\label{tab:mechanisms}
\centering
\footnotesize
\setlength{\tabcolsep}{3pt}
\begin{tabular}{@{}llll@{}}
\toprule
method & prediction & executed & routing reads \\
\midrule
\cp & -- & router+bonus & LRU state \\
SpecMD (bonus) & next router & router+bonus & physical \\
SMoE, MoE-CORE & next router & router, subst. & physical \\
Spec.\ Experts & next router$^*$ & predicted & none \\
CommitMoE & trained MLP & predicted & none \\
\psr & next router & router+bonus & logical \\
\bottomrule
\end{tabular}
\\[2pt]{\footnotesize $^*$with a default vector or a trained estimator.}
\end{table}

\textbf{Training for locality.} Several works fine-tune or train routers toward expert reuse: ReMoE~\citep{zhu2026remoe}, Sticky Routing~\citep{kayyam2026sticky}, temporally extended MoE~\citep{shen2026temoe}, MELINOE~\citep{raje2026melinoe}, cache-aware joint router adaptation~\citep{wu2026cacheaware}, and segment-level routing~\citep{kundu2026stepped}. In a pre-registered study, \citet{suram2026cacheable} train small MoE models from scratch with locality losses and find that the reduction in cache misses cannot be had without a perplexity cost. Our \cref{sec:training} adds a paired-benchmark comparison against training-free cache-aware routing.

\textbf{Speculative decoding of offloaded MoE models.} A verification step needs the union of its positions' experts. Under offloading, AcceptMoE~\citep{liang2026acceptmoe} finds standard speculative decoding slower than AR decoding on 7 of the 8 model pairs it tests, and \citet{saxena2025utility} measure 2--3$\times$ more data per verification and switch speculation off when it does not pay. Prefetching experts during drafting~\citep{chen2025spmoe,wang2025moespeq,han2026draftexpert} gives copies more lead time but does not reduce them. Restricting the verifier's experts reduces them at a quality cost~\citep{mcdanel2026moespec,liang2026acceptmoe,huang2026s2moe}. We evaluate AcceptMoE's block-level selection with drafter-confidence weights and a top-1 rescue in an offloading engine against \psr, including the drafter's cost (\cref{sec:sd}). In the methods that restrict the verifier's experts, the routing of a committed token depends on the whole draft block; \cref{sec:sd} measures how much.

\textbf{Evaluation methodology.} \citet{zhang2026reproducible} discuss replay semantics and workload contamination in trace-based evaluation of expert caching. We address a different issue: when routing depends on the cache, the cache becomes part of the evaluated model. A deterministic logical cache provides timing-independent routing semantics that an evaluation harness can reproduce; the alternatives of \cref{sec:semantics}, benchmarking inside the deployed engine or fixing the order of copies and computation, give results for one hardware and load configuration.

\section{Limitations}
\label{sec:limits}

\textbf{Three models, one GPU.} OLMoE fits on the GPU we use, so its memory pressure is emulated; for Moonlight and Qwen3.6 it is real. OLMoE and Moonlight have 64 fine-grained experts per layer, Qwen3.6 has 256; models with fewer, larger experts (e.g.\ Mixtral) leave less room for cache-aware routing. The qualitative findings transferred from OLMoE to Moonlight without retuning except for the pre-gate window, which depends on how close the link is to saturation. They did not fully transfer to Qwen3.6: there copies are short relative to computation, \psr's latency advantage over \cp disappears, and only its lower divergence and its gains on the likelihood benchmarks remain (\cref{sec:qwen}). The numbers need to be re-measured for other models and links. Moonlight's and Qwen3.6's benchmarks were decoded in batches, and Qwen3.6 was evaluated on six tasks with a changed GSM8K protocol (\cref{app:protocol}).

\textbf{No external systems.} Our baselines are reimplementations in one engine, which isolates the policies but does not show how \psr compares with tuned systems such as llama.cpp, MoE-Infinity, or CPU-execution systems~\citep{kamahori2024fiddler,chen2025ktransformers}; at batch size one, computing missing experts on a strong CPU is a competitive alternative that \psr could be combined with. We did not implement SMoE-style substitution.

\textbf{Measurement.} The main latency results are teacher-forced over a fixed text set and exclude prompt processing. In greedy generation on benchmark prompts the advantage of \psr over \cp is smaller and reverses on HumanEval (\cref{sec:engine-quality}): it depends on how much expert locality the decoded text has, and short, repetitive outputs leave little for prefetching to win. Benchmarks use greedy decoding; paired standard errors on the generative tasks are 0.8--3 points, so differences below that are not resolved. Host-side selection is part of the measured latency: in the OLMoE engine \cp and \psr pay about 0.3\,ms per token for it and exact routing does not, which makes their speedups over exact routing slightly conservative; on Qwen3.6 an uncompiled selection cost 4.6\,ms per token, and we report the compiled one (\cref{sec:qwen}).

\textbf{Numerics.} The harness and the engine agree up to bf16 near-ties, which a cache-conditioned router propagates; this is the same kind of discrepancy any two implementations of exact routing have. Benchmark comparisons can therefore be checked against the engine only statistically, not token for token: under exact routing \ebDiscExactGsm\% of GSM8K answers differ between the two. The paired differences between configurations agree within the resolution of that check ($\pm$\ebCImin--\ebCImax\ points, \cref{sec:engine-quality}), which excludes large discrepancies but not ones the size of most per-task differences we report.

\textbf{Speculative decoding.} \Cref{sec:sd} uses greedy decoding and one drafter per target. On OLMoE its conclusion that a much cheaper drafter would be needed rests on measured components. The block drafter on Qwen3.6 confirms it for latency, but the quality of block selection there rests on 64 prompts, and its latency advantage is measured for our kernels and these prompts.

\section{Conclusion}

Cache-aware routing makes the cache part of the model. A deterministic logical cache makes the routing policy independent of transfer timing, so an evaluation harness can reproduce it. On that basis, letting a small budget of cross-layer prefetches choose which experts receive the routing bonus gives the best latency--quality trade-off we measured: \spdExactLruRng$\times$ faster than exact routing with LRU, and \redCPA--\redCPB\% lower latency than the \cp operating point with mean benchmark differences whose 95\% lower bounds are \pairIsoQualAlo\ and \pairIsoQualBlo\ points, on OLMoE with half or a quarter of its experts resident. These benchmark comparisons come from the evaluation harness; benchmarks generated by the engine itself found no discrepancy beyond that check's resolution of \ebCImin--\ebCImax\ points per task, which rules out a large bias but does not show that per-task differences of one or two points carry over to the engine. On Moonlight-16B-A3B, which must be offloaded, \psr is \spdMoonExact$\times$ faster than exact routing with LRU at 37.5\% residency and, at the latency of \cp, better on three of eight benchmarks after Holm's correction and worse on none. On Qwen3.6-35B-A3B, where copies are cheap relative to computation, \psr keeps its lower divergence and its gains on MMLU and HellaSwag at the latency of \cp, but not its speed: its latency benefit is specific to transfer-bound decoding. Speculative decoding with block-level expert selection saves expert bytes when drafts are accepted in long runs; with a 1B autoregressive drafter on OLMoE that saving did not pay, while with a block drafter on Qwen3.6 it made decoding \spdQwSdCPRng$\times$ faster than any autoregressive configuration with at most its GPU memory on the prompts we timed.

\bibliography{refs}
\bibliographystyle{icml2026}

\newpage
\appendix
\onecolumn

\section{Benchmark protocol details}
\label{app:protocol}

All benchmarks use lm-evaluation-harness 0.4.13 with transformers 5.12 and PyTorch 2.11, with the routing policy applied in every MoE layer for every prompt and generated token. OLMoE runs at batch size 1 with bf16 weights; Moonlight and Qwen3.6 (FP8 weights) are batched as described below. Policy state (logical cache, moving averages) is reset at the start of each request. MMLU, ARC-Challenge and HellaSwag report accuracy (not length-normalized accuracy). GSM8K uses 5-shot prompts and strict-match extraction. BBH uses the 3-shot chain-of-thought prompts with the answer-extraction filter, limited to the first 100 questions of each of the 27 subtasks. IFEval uses the model's chat template. HumanEval and MBPP execute the generated code; with greedy decoding pass@1 is a 0/1 outcome per problem. MMLU subtasks and BBH subtasks are pooled at the question level. For IFEval at instruction level the prompt is the unit of every paired statistic (cluster-robust standard error, sign-flip test on per-prompt sums; \cref{sec:setup}) on all models and in the engine check.

\textbf{Moonlight.} Streaming experts from host memory at batch 1 (about 85\,ms per token) is too slow for the generative benchmarks, so we hold the whole model on two GPUs, which does not change routing because routing depends only on the logical cache, and decode in batches of 8 with left padding (MBPP in batches of 4, because of one 2736-token prompt). Each row of a batch keeps its own logical cache and statistics, and padded positions do not update them. Replaying the per-row cache update on recorded router outputs reproduces the batched selections exactly. Batched and batch-1 forward passes differ only by bf16 near-ties: 94\% of prompt token-layers select identical expert sets, the same rate as between the harness's own incremental and full forward passes at batch 1 (94.9\%). All configurations use the same batch size per task.

\textbf{Qwen3.6.} The harness holds the FP8 model on two GPUs (vLLM's block-FP8 kernels) and decodes in batches of 8 with left padding and per-row logical caches, as for Moonlight. The likelihood tasks run in batches of 2, as the log-softmax over the 248k-token vocabulary does not fit in memory for 8. We ran six of the eight tasks (no BBH or MBPP). The chat template is used with the model's reasoning mode disabled. On GSM8K's plain-text 5-shot prompt, Qwen3.6 often opens a reasoning block anyway, whose text hits the task's stop string (``Question:'') after a few tokens; GSM8K therefore uses the chat template with the five shots as conversation turns and at most 512 new tokens (at 256, 4 of 32 answers were cut in a pilot). IFEval uses the chat template and HumanEval the plain completion prompt. All configurations use the same protocol and batch sizes. Engine timing and speculative decoding use one GPU with the routed experts in pinned host memory.

\textbf{Speculative decoding.} The scores of \cref{tab:sd} come from a speculative decoding loop in the HuggingFace harness with greedy drafts and verification, which implements block selection inside the target's forward pass with the engine's selection and pool; task protocols are as above.

\textbf{Training (\cref{sec:training}).} Both studies distil the frozen model on a token set disjoint from all evaluation data: OLMoE's own answers to held-out WildChat prompts (the split used for the token prior), UltraChat test conversations, and C4 training text, each document at most 1024 tokens. AdamW ($\beta=(0.9,0.95)$, no weight decay), learning rate $10^{-4}$ for the LoRA adapters and margin head and $2\cdot10^{-5}$ for the router, 20 warm-up steps, data-parallel on 6 GPUs with gradient accumulation; seed 0, with a seed-1 replicate of two gate variants. The objective is the KL divergence to the exact model's next-token distribution plus small penalties on host loads and rapid set changes. The search covered 27 runs for the change-based gate and 18 for post-training \psr (including early-stopped variants), benchmarked after screening on selfgen and copies.

\section{Configuration names}
\label{app:configs}

\psr $\lambda$, $n$: \cp bonus $\lambda$ on the logical cache, A priority, pre-gate candidates 8, at most $n$ new insertions per layer and token, at most two prefetch copies in flight. The token prior, the bonus scale $\bar\Delta_l$ and the warm set (the $C$ most frequently selected experts per layer) are estimated on a held-out split of the training-side text corpus that excludes all evaluation data. On Moonlight the pre-gate candidates are the first 4 of the ranking (\cref{sec:moonlight}); on Qwen3.6 the first 4 or 2 of 16 (\cref{sec:qwen}). \cp with top-$J$ protection gives the router's top-$J$ experts the bonus as if they were resident.

Block-level selection (\cref{sec:sd}): per layer and verification block, the eligible set is the root position's own top-$K$, $A$, plus the $\min(\lceil\gamma e^{H}\rceil, m)$ experts outside $A$ with the highest weighted demand $u_e=\sum_d\alpha_d\,r_{d,e}\,\mathbb{1}[e\in\mathrm{TopK}(z_d)]$, where $r_d$ is the router softmax, $m$ the number of experts outside $A$ with $u_e>0$, and $H=-\sum_{e\notin A}q_e\ln q_e$ the entropy, in nats, of the normalized demand $q_e=u_e/\sum_{e'\notin A}u_{e'}$. Non-resident members are then pruned in increasing order of demand within AcceptMoE's rerouting budget~\citep{liang2026acceptmoe}. Weights: position weights $\alpha_d=p_d^{1/2}$ with $p_d$ the measured commitment rate at depth $d$ and $\gamma=1$ (AcceptMoE); confidence weights $\alpha_d=(\prod_{i\le d}c_i)^{1/2}$ with $\gamma=1.05$; $\alpha_0=1$. The pool holds $P=24$ experts per layer on OLMoE (64 on Qwen3.6) and is refilled by recency after each block.

\end{document}